\documentclass[journal]{IEEEtran}

\ifCLASSINFOpdf
  \usepackage[pdftex]{graphicx}
  \graphicspath{{../pdf/}{../jpeg/}}
  \DeclareGraphicsExtensions{.pdf,.jpeg,.png}
\else
  \usepackage[dvips]{graphicx}
  \graphicspath{{../eps/}}
  \DeclareGraphicsExtensions{.eps}
\fi

\usepackage{cite}
\usepackage{graphicx}
\usepackage{subcaption}
\usepackage{amssymb}
\usepackage{hyperref}
\hypersetup{colorlinks, citecolor=green, filecolor=pink, linkcolor=red, urlcolor=blue }
\begin{document}
%
\title{Artificial Intelligence-Driven Customized Manufacturing Factory: Key Technologies, Applications, and Challenges}
%
%
%

\author{Jiafu~Wan,~\IEEEmembership{Member,~IEEE,}
        Xiaomin~Li,
        Hong-Ning~Dai,~\IEEEmembership{Senior Member,~IEEE,}
        Andrew~Kusiak,~\IEEEmembership{Life Member,~IEEE,}
        Miguel Mart\'{i}nez-Garc\'{i}a,~\IEEEmembership{Member,~IEEE,}
        and~Di Li
\thanks{This work was supported in part by the National Key R \& D Program of China (Grant No. 2018YFB1700500), the Joint Fund of the National Natural Science Foundation of China and Guangdong Province (Grant No. U1801264), and Macao Science and Technology Development Fund under Macao Funding Scheme for Key R \& D Projects (0025/2019/AKP). ({\it Corresponding author: Hong-Ning Dai.})}
\thanks{J. Wan and D. Li are with the School of Mechanical and Automotive Engineering, South China University of Technology, Guangzhou, China (e-mails: mejwan@scut.edu.cn, itdili@scut.edu.cn).}
\thanks{X. Li is with the School of Mechanical Engineering, Zhongkai University of Agriculture and Engineering, Guangzhou, China (e-mail: lixiaomin@zhku.edu.cn).}
\thanks{H.-N. Dai is with the Faculty of Information Technology, Macau University of Science and Technology, Macau SAR (email: hndai@ieee.org).}
\thanks{A. Kusiak is with the Intelligent Systems Laboratory, Department of Mechanical and Industrial Engineering, The University of Iowa, Iowa City, USA (email: andrew-kusiak@uiowa.edu).}
\thanks{M. Mart\'{i}nez-Garc\'{i}a is with the Dept. of Aeronautical and Automotive Engineering, Loughborough University, UK (email: m.martinez-garcia@lboro.ac.uk).}
\thanks{Manuscript received xx; revised xx.}}

%
%

\markboth{Proceedings of the IEEE, February~2023}%
{Shell \MakeLowercase{\textit{et al.}}: Artificial Intelligence-Driven Customized Manufacturing Factory: Key Technologies, Applications, and Challenges}
%



\maketitle

\begin{abstract}
The traditional production paradigm of large batch production does not offer flexibility towards satisfying the requirements of individual customers. A new generation of smart factories is expected to support new multi-variety and small-batch customized production modes. For that, Artificial Intelligence (AI) is enabling higher value-added manufacturing by accelerating the integration of manufacturing and information communication technologies, including computing, communication, and control. The characteristics of a customized smart factory are to include self-perception, operations optimization, dynamic reconfiguration, and intelligent decision-making. The AI technologies will allow manufacturing systems to perceive the environment, adapt to external needs, and extract the processing knowledge, including business models, such as intelligent production, networked collaboration, and extended service models.

This paper focuses on the implementation of AI in customized manufacturing (CM). The architecture of an AI-driven customized smart factory is presented. Details of intelligent manufacturing devices, intelligent information interaction, and the construction of a flexible manufacturing line are showcased. The state-of-the-art AI technologies of potential use in CM, i.e., machine learning, multi-agent systems, Internet of Things, big data, and cloud-edge computing are surveyed. The AI-enabled technologies in a customized smart factory are validated with a case study of customized packaging. The experimental results have demonstrated that the AI-assisted CM offers the possibility of higher production flexibility and efficiency. Challenges and solutions related to AI in CM are also discussed.

\end{abstract}

\begin{IEEEkeywords}
Customized Manufacturing; Artificial Intelligence; Industry 4.0; Smart Factory; Software-Defined Network. 
\end{IEEEkeywords}

%
\IEEEpeerreviewmaketitle

\section{Introduction}
%
%
%
%
\IEEEPARstart{T}{he} Industry 4.0 initiative is advocating smart manufacturing as the industrial revolution leading to global economic growth~\cite{1,2,3,4}. Many  countries, corporations, and research institutions have embraced the concept of Industry 4.0, in particular the United States, the European Union, and East Asia~\cite{5}. Some industries have begun a transformation from the digital era to the intelligent era. Manufacturing represents a large segment of the global economy, while the interest in smart manufacturing is expanding~\cite{6}. The progress in information and communication technologies, for example, the Internet of Things (IoT)~\cite{7,8}, artificial intelligence (AI)~\cite{9,10}, AI-Generated Content (AIGC)~\cite{cao2023comprehensive}, and big data~\cite{11,12} for manufacturing applications, has impacted smart manufacturing~\cite{13}. In the broad context of manufacturing, customized manufacturing (CM) offers a value-added paradigm for smart manufacturing~\cite{14}, as it refers to personalized products and services. The benefits of CM have been highlighted by multinational companies.

Today, information and communication technologies are the base of smart manufacturing~\cite{15, 16}, and intelligent systems driven by AI are the core of CM~\cite{17}. With the development of AI technologies, new theories, models, algorithms, and applications - towards simulating, extending, and enhancing human intelligence - are continuously developed. The progress of big data analysis and deep learning has accelerated AI to enter the 2.0 era~\cite{18,19,20}. AI 2.0 manifests itself as a data-driven deep reinforcement learning intelligence~\cite{21}, network-based swarm intelligence~\cite{22}, technology-oriented hybrid intelligence of human-machine and brain-machine interaction~\cite{Martinez-Garcia2018,23,martinez2019memory}, cross-media reasoning intelligence~\cite{24, 25}, etc. Therefore, AI 2.0 offers significant potential to smart manufacturing, especially, CM in smart factories~\cite{26}.

Typically, AI solutions can be applied to several aspects of smart manufacturing. AI algorithms can run the manufacturing of personalized products in a smart factory~\cite{27, 28}. The AI-assisted CM is to construct smart manufacturing systems supported by cognitive computing, machine status sensing, real-time data analysis, and autonomous decision-making~\cite{29, 30}. AI permeates through every link of CM value chains, such as design, production, management, and service~\cite{31, 32}. Based on these insights of CM and AI, the focus of this paper is on the implementation of AI in the smart factory for CM involving architecture, manufacturing equipment, information exchange, flexible production line, and smart manufacturing services.

The contributions of the research presented in this paper are as follows.

\begin{itemize}

\item The architecture of the AI-assisted CM for smart factories is developed by merging smart devices and industrial networks with big data analysis.

\item The state-of-the-art AI technologies are reviewed and discussed.
\item The key AI-enabled technologies in CM are validated with a prototype platform of a customized candy packaging line.
\item The challenges and possible solutions brought by the introduction of AI into CM are discussed.

\end{itemize}

The remainder of the paper is organized as follows. In Section~\ref{sec:cm-ai}, the relationship between the CM and AI is discussed. The general architecture of AI-assisted CM is presented in Section~\ref{sec:arch}. Section~\ref{sec:devices} illustrates the implementation of AI in intelligent manufacturing equipment. The intelligent information exchange process, flexible production line, and smart manufacturing services in the AI-assisted CM are proposed in Section~\ref{sec:interaction} and Section~\ref{sec:manu-line}, respectively. A case study is provided in Section~\ref{sec:case}. The challenges and possible solutions to the AI-assisted intelligent manufacturing factory are discussed in Section~\ref{sec:challenges}. Section~\ref{sec:conc} concludes the paper.

\section{Customized Manufacturing and Artificial Intelligence }
\label{sec:cm-ai}

This section first summarizes the characteristics of customized manufacturing in Section~\ref{subsec:Char-CM} and then discusses the opportunities brought by AI-driven customized manufacturing in Section~\ref{subsec:AI-CM}.

\subsection{Characteristics of customized manufacturing}
\label{subsec:Char-CM}

Despite the progress made, manufacturing industry faces a number of challenges, some of which are: traditional mass-production is not able to adapt to the rapid production of personalized products; and resource limitations, environmental pollution, global warming, and an aging global population have become more prominent. Therefore, a new manufacturing paradigm to address these challenges is needed. The customer-to-manufacture concept reflects the characteristics of customized production where a manufacturing system directly interacts with a customer to meet his/her personalized needs. The goal is to realize the rapid customization of personalized products. The new generation of intelligent manufacturing technology offers improved flexibility, transparency, resource utilization, and efficiency of manufacturing processes. It has led to new programs, e.g., the Factory of the Future in Europe~\cite{33}, Industry 4.0 in Germany~\cite{1}, and Made in China 2025~\cite{34}. Moreover, the United States has accelerated research and development programs~\cite{35}.

Compared with mass production, the production organization of CM is more complex, quality control is more difficult, and the energy consumption needs attention. In classical automation, the production boundaries were rigid to ensure quality, cost, and efficiency. Compared with traditional production, CM has the following characteristics.

\begin{itemize}
    \item \emph{Smart interconnectivity.} Smart manufacturing embraces a cyber-physical environment, e.g., processing/detection/assembly equipment, and storage, all operating in a heterogeneous industrial network. The Industrial IoT has progressed from the original industrial sensor networks to the Narrow Band-Internet of Things (NB-IoT), LoRa WAN, and LTE Cat M1 with increased coverage at reduced power consumption~\cite{36}. Edge computing units are deployed to improve system intelligence. Cognitive technology ensures the context awareness and semantic understanding of the industrial IoT~\cite{37}. Intelligent industrial IoT as the key technologies is widely used for intelligent manufacturing.
    \item \emph{Dynamic reconfiguration.} The concept of a smart factory aims at the rapid manufacturing of a variety of products in small batches. Since the product types may change dynamically, system resources need to be dynamically reorganized. A multi-agent system~\cite{38} is introduced to negotiate a new system configuration.
    \item \emph{Massive volumes of data.} An intelligent manufacturing system includes interconnected devices generating data such as device status and process parameters. Cloud computing and big data science make data analysis feasible in failure prediction, active preventive maintenance, and decision making.
    \item \emph{Deep integration.} The underlying intelligent manufacturing entities, cloud platforms, edge servers, and upper monitoring terminals are closely connected. Data processing, control, and operations can be performed simultaneously in the Cyber-Physical Systems (CPS), where the information barriers are broken down, thereby realizing the deep integration of physical and information environments.
\end{itemize}


\begin{figure*}[h]
\centering
\begin{subfigure}[t]{0.45\textwidth}
\includegraphics[height=6.3cm]{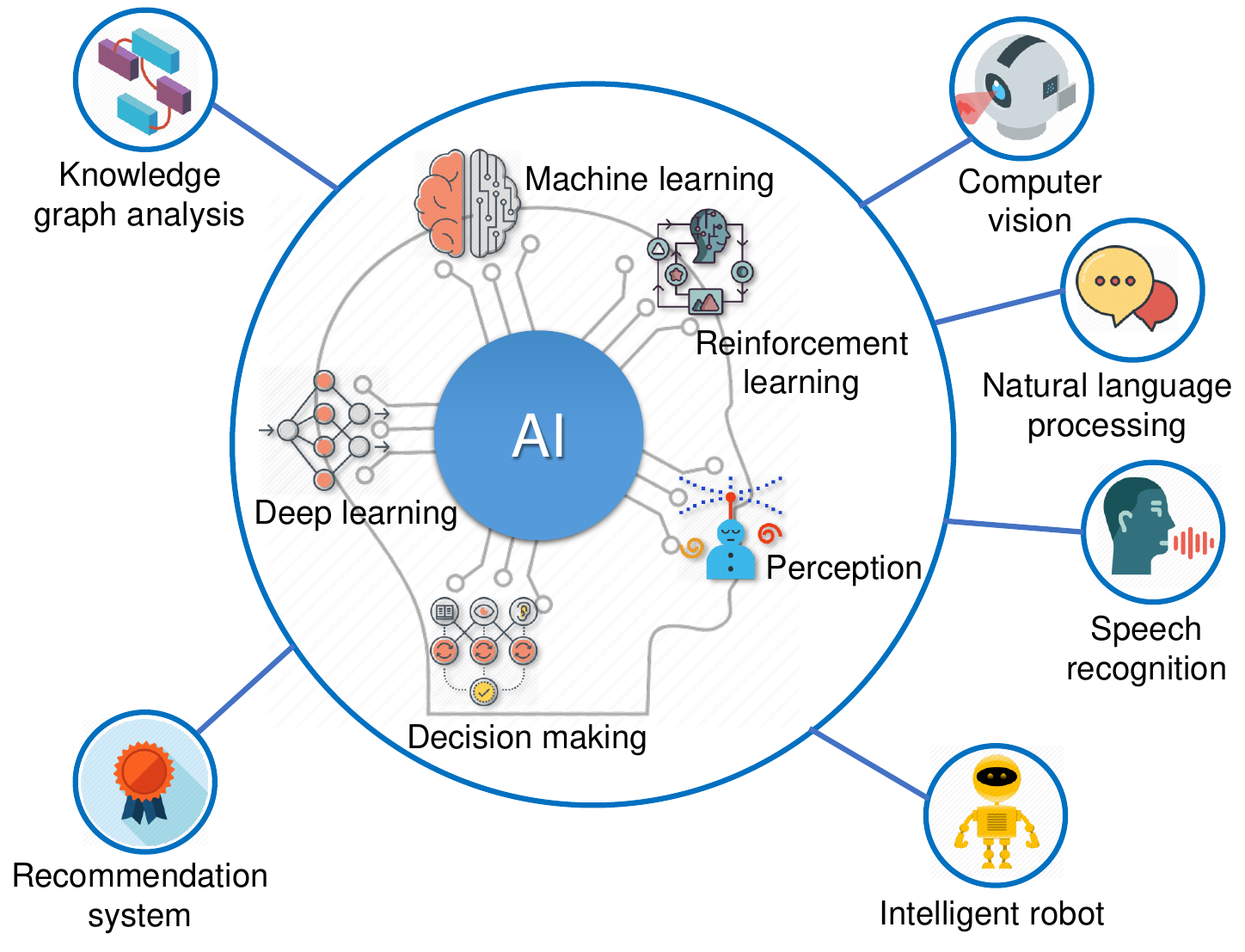}
\caption{}
\label{fig:fig1a}
\end{subfigure}
\begin{subfigure}[t]{0.45\textwidth}
\includegraphics[height=6.3cm]{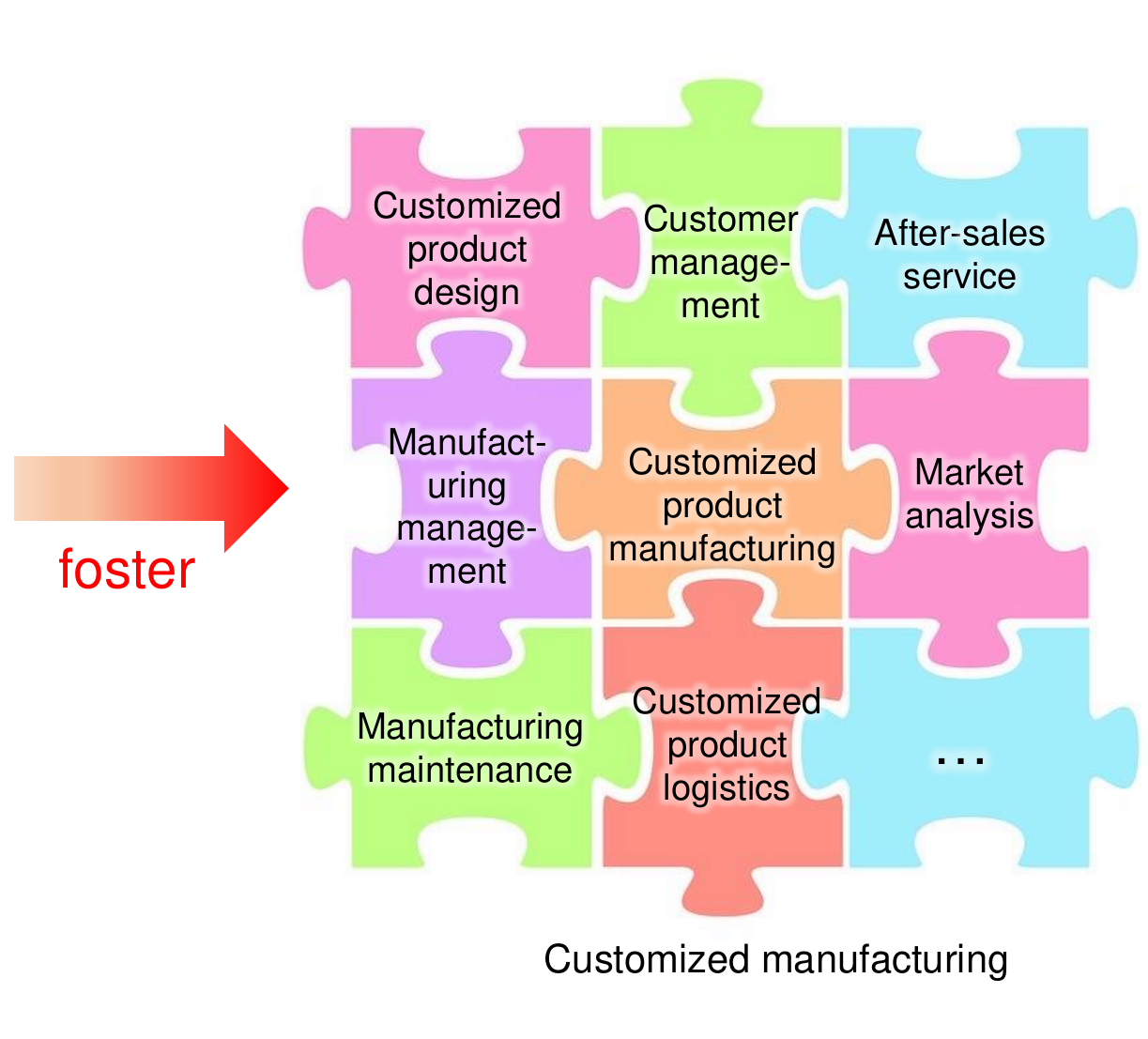}
\caption{}
\label{fig:fig1b}
\end{subfigure}%
\caption{The AI and customized manufacturing. (a) AI technologies include perception, machine learning, deep learning, reinforcement learning, and decision making as well as AI-enabled applications like computer vision, natural language processing, intelligent robots, and recommendation systems. (b) AI can foster customized manufacturing in the aspects: customized product design, customized product manufacturing, manufacturing maintenance, customer management, logistics, after-sales service, and market analysis.}
\label{figure1}
\end{figure*}

\subsection{Overview of AI technologies}

AI embraces theories, methods, technologies, and applications to augment human intelligence. It includes not only AI techniques such as perception, machine learning (ML), deep learning (DL), reinforcement learning, and decision making, but also AI-enabled applications like computer vision, natural language processing, intelligent robots, and recommendation systems, as shown in Fig.~\ref{fig:fig1a}. ML has outperformed traditional statistical methods in tasks such as classification, regression, clustering, and rule extraction~\cite{hndai:EIS19}. Typical ML algorithms include decision tree, support vector machines, regression analysis, Bayesian networks, and deep neural networks. Recent advanced AI approaches like AIGC~\cite{cao2023comprehensive} can be used to intelligently direct the interaction between customers and product designers.

As a subset of ML algorithms, DL algorithms have superior performance than other ML algorithms. The recent success of DL algorithms mainly owes to three factors: 1) the availability of massive data; 2) the advent of computer capability achieved by computer architectures and hardware, such as Graphic Processing Units (GPUs); 3) the advances in diverse DL algorithms such as a convolutional neural network (CNN), long short-term memory (LSTM) and their variants. Different from ML methods, which require substantial efforts in feature engineering in processing raw industrial data, DL methods combine feature engineering and learning process together, thereby achieving outstanding performance. 

However, DL algorithms also have their disadvantages. First, DL algorithms often require a huge amount of data to train DL models to achieve better performance than other ML algorithms. Moreover, the training of DL models requires substantial computing resources (e.g., expensive GPUs and other computer hardware devices). Third, DL algorithms also suffer from poor interpretability, i.e., a DL model is like an uncontrollable ``black box'', which may not obtain the result as predicted. The poor interpretability of DL models may prevent their wide adoption in industrial systems, especially in critical tasks like fault diagnosis~\cite{HWang:TII20} despite recent advances in improving the interpretability of DL models~\cite{XZhang:TII20}. 

\subsection{AI-driven customized manufacturing}
\label{subsec:AI-CM}

\begin{figure*}[!t]
  \centering
   \includegraphics[scale= 0.7]{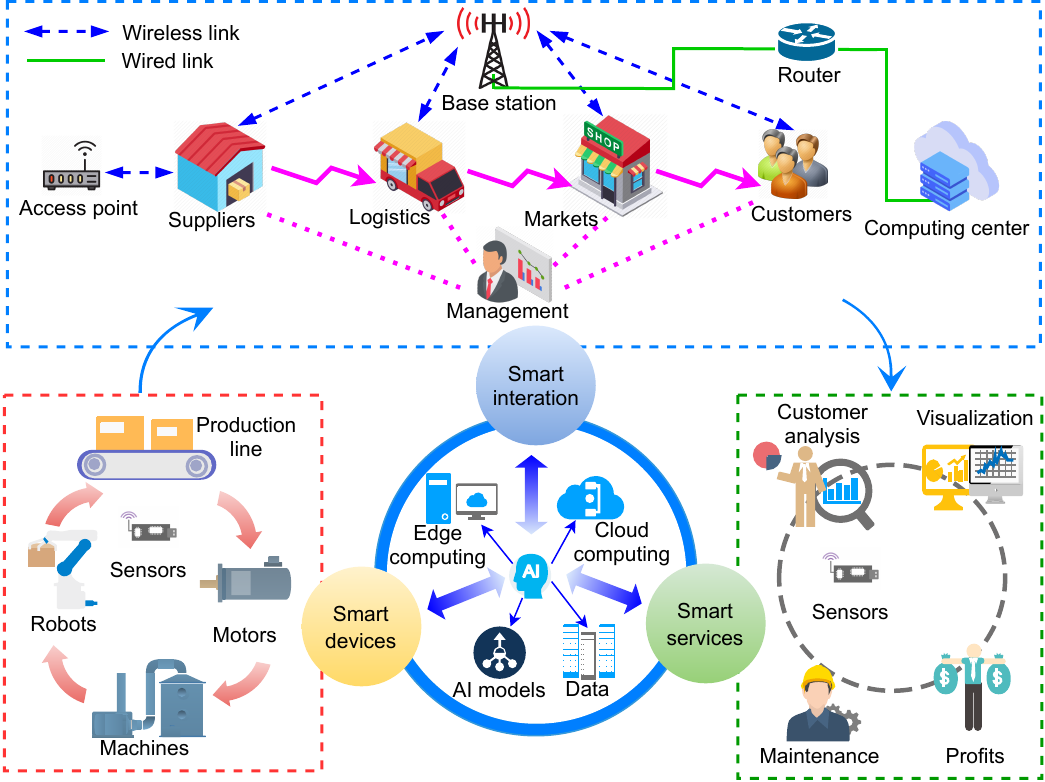}
        \caption{The architecture of AI-assisted customized manufacturing includes smart devices, smart interaction, AI layer, and smart services.}
    \label{figure2}
\end{figure*}

As AI technologies have demonstrated their potential in areas such as customized product design, customized product manufacturing, manufacturing management, manufacturing maintenance, customer management, logistics, after-sales service, and market analysis as shown in Fig.~\ref{fig:fig1b}, industrial practitioners and researchers have begun their implementation. For example, the work~\cite{zuo:2016prediction} presents a Bayesian network-based approach to analyze the consumers' purchase behaviour via analyzing RFID data, which is collected from RFID-tags attached to in-store shopping carts. Moreover, a deep learning method is adopted to identify possible machine faults through analyzing mechanic data collected from the real industrial environments such as induction motors, gearboxes, and bearings~\cite{SShao:TII19}.

Therefore, the introduction of AI technologies can potentially realize the customized manufacturing. We name such AI-driven customized manufacturing as AI-driven CM. In summary, AI-driven CM has the following advantages~\cite{39,40}.

\begin{enumerate}

\item \emph{Improved production efficiency and product quality.}
In CM factories, automated devices can potentially make decisions with reduced human interventions. Technologies such as ML and computer vision are enablers of cognitive capabilities, learning, and reasoning (e.g., analysis of order quantities, lead time, faults, errors, and downtime). Product defects and process anomalies can be identified using computer vision and foreign object detection. Human operators can be alerted to process deviations.

\item \emph{Facilitating predictive maintenance.}
Scheduled maintenance ensures that the equipment is in the best state. Sensors installed on a production line collect data for analysis with ML algorithms, including convolutional neural networks. For example, the wear and tear of a machine can be detected in real-time and a notification can be issued.

\item \emph{Developing of smart supply chains.}
The variability and uncertainty of supply chains for CM can be predicted with ML algorithms. Moreover, the insights obtained can be used to predict sudden changes in customer demands.

\end{enumerate}

In short, the incorporation of AI and industrial IoT brings benefits to smart manufacturing. AI-assisted tools improve manufacturing efficiency. Meanwhile, higher value-added products can be introduced to the market. 

However, we cannot deny that AI technologies still have their limitations when they are formally adopted to real-world manufacturing scenarios. On the one hand, AI and ML algorithms often have stringent requirements on computing facilities. For example, high-performance computing servers equipped with GPUs are often required to fasten the training process on massive data~\cite{VSze:PIEEE17} while exiting manufacturing facilities may not fulfill the stringent requirement on computing capability. Therefore, the common practice is to outsource (or upload) the manufacturing data to cloud computing service providers who can conduct the computing-intensive tasks. Nevertheless, outsourcing the manufacturing data to the third party may lead to the risk of leaking confidential data (e.g., customized product design) or exposing private customer data to others. On the other hand, transferring the manufacturing data to remote clouds inevitably leads to high latency, thereby failing to fulfill the real-time requirement of time-sensitive tasks.

\section{Architecture of an AI-Assisted Customized Manufacturing Factory}
\label{sec:arch}

This section first presents an AI-Assisted customized manufacturing (AIaCM) framework in Section~\ref{subsec:AIaCM} and then gives a brief comparison of the proposed AIaCM framework with the state-of-the-art literature in Section~\ref{subsec:survey}. 

\subsection{AI-Assisted Customized Manufacturing Factory}
\label{subsec:AIaCM}
Different frameworks have been presented towards the increased interactivity and resource management~\cite{41,42,43}. Most studies have focused on information communications~\cite{44} or big data processing~\cite{45,46,47}. So far, research proposing generic AI-based CM frameworks is limited. System performance metrics, e.g., flexibility, efficiency, scalability, and sustainability, can be improved by adopting AI technologies such as ML, knowledge graphs, and human-computer interaction (HCI). This is especially true in sensing, interaction, resource optimization, operations, and maintenance in a smart CM factory~\cite{48, 49}. Since cloud computing, edge computing, and local computing paradigms have their unique strengths and limitations, they should be integrated to maximize their effectiveness. At the same time, the corresponding AI algorithms should be redesigned to match the corresponding computing paradigm. Cloud intelligence is responsible for making comprehensive, time-insensitive analysis and decisions, while the edge and local node intelligence are applicable to the context or time-aware environments. Intelligent manufacturing systems include smart manufacturing devices, realize intelligent information interaction, and provide intelligent manufacturing services by merging AI technologies. As shown in Fig.~\ref{figure2}, an AI-assisted CM framework that includes smart devices, smart interaction, AI layer, and smart services. We then explain this framework in detail as follows.

\subsubsection{Smart devices} include robots, conveyors, and other basic controlled platforms. Smart devices serve as ``the physical layer'' for the entire AIaCM. Specifically, different devices and equipment, such as robots and processing tools are controlled by their corresponding automatic control systems. Therefore, it is crucial to meet the real-time requirement for the device layer in an AIaCM system. To achieve this goal, ML algorithms can be implemented at the device layer in low power devices such as FPGAs. The interconnection of the physical devices, e.g., machines, conveyors, is implemented at the device layer~\cite{50,51} using edge computing servers. 

\begin{table*}[t]
\centering
\caption{Summary of most relevant state-of-the-art literature}
\renewcommand{\arraystretch}{1.5}
\label{tab:comp}

\begin{tabular}{|c|c|c|c|c||p{2.7cm}|p{2.7cm}|p{2.8cm}|}
\hline
\textbf{Refs.} & \textbf{Smart devices} & \textbf{Smart interaction} & \textbf{AI} & \textbf{Smart services} & \textbf{Pros} & \textbf{Cons} & \textbf{Applications} \\ \hline\hline
  \cite{41}    &     \checkmark      &     $\times$              & \checkmark   &  \checkmark      &    Integration sensor with cloud services  & No edge computing considered     &  Machine status monitoring (primitive ML methods were used)          \\ \hline
  \cite{42}    &        \checkmark       &      \checkmark             &  \checkmark  &      \checkmark      &  Service-oriented smart
manufacturing   &   No edge computing considered   &    Milk production from buffalo pasture          \\ \hline
  \cite{43}    &   \checkmark    &   \checkmark         &  $\times$   &    $\times$    &   Integration CPS with smart manufacturing   &  No edge computing and in-depth analysis of AI algorithms    &     Several cases from product design to manufacturing control \\ \hline
  \cite{44}    &     \checkmark          &     $\times$              &  $\times$  &     \checkmark        & Comprehensive consideration of the entire industrial network     &   No AI as well as edge computing considered    &    No specific application          \\ \hline
  \cite{46}    &    \checkmark          &    $\times$                &  \checkmark  &    \checkmark   &     Integration sensor with cloud services  &   No edge computing considered   &    Light gauge steel production line      \\ \hline
  \cite{47}   &    \checkmark          &    \checkmark                  &  \checkmark  &    $\times$   &     Integration CPS with AI  &   No edge computing considered   &   Production line and factory management    \\ \hline
  \cite{48}    &      $\times$          &     $\times$              &   \checkmark &    $\times$           & Diverse AI algorithms were used &  No consideration of smart devices, interactions and services &    Cold spray additive manufacturing,  augmented reality-guided inspection and surface stress estimation    \\ \hline
\end{tabular}
\end{table*}

\subsubsection{Smart interaction}links the device layer, AI layer, and services layer~\cite{52,53}. It represents a bridge between different layers of the proposed architecture. The smart interaction layer is composed of two vital modules. The first module includes basic network devices such as access points, switches, routers and network controllers, which are generally supported by different network operating systems, or equipped with different network functions. The basic network devices constitute the core of the network layer~\cite{54,55}. Different from the first module which is fixed or static, the second module consists of the dynamic elements, including network/communications protocols, information interaction, and data persistent or transient storage. These dynamic elements are essentially information carriers to connect different manufacturing processes. The dynamic module is running on top of the static one.

AI is utilized in the prediction of wireless channels, optimization of mobile network handoffs, and control network congestion. Recurrent Neural Networks (RNN) or Reservoir Computing (RC) are candidate solutions due to their advantages of them in analyzing temporal network data.

\subsubsection {The AI layer} includes algorithms running at different computing platforms such as edge or cloud servers~\cite{46,56}. The computing environment consists of cloud and edge computing servers running MapReduce, Hadoop, and Spark.

AI algorithms are adopted at different levels of computing paradigms in the AIaCM architecture. For instance, training a deep learning model for image processing can be conducted in the cloud. Then, edge computing servers are responsible for running the trained DL model and executing relatively simple algorithms for specific manufacturing tasks.

\subsubsection{Smart manufacturing services} include data visualization, system maintenance, predictions, and market analysis. For example, a recommender system can provide customers with details of CM products, and the information including the performance of a production line, market trends, and efficiency of the supply chain. 

\subsection{Overview on state-of-the-art manufacturing methods}
\label{subsec:survey}

Recently, substantial research efforts have been made to improve the interactivity and elasticity of exiting manufacturing factories~\cite{41,42,43,44,45,46,47,48,49}. Table~\ref{tab:comp} summarizes most relevant state-of-the-art literature. We can observe from  Table~\ref{tab:comp} that most of the references only concentrate on a single or several aspects in CM. For example, the work~\cite{41} presents a cloud manufacturing framework to analyze and process manufacturing data. Similarly, a cloud-based manufacturing equipment~\cite{46} is proposed to provide users with on-demand services. However, outsourcing manufacturing data to cloud services providers who are often owned by third parties can also bring the risks of leaking customers' private data and exposing confidential manufacturing data (e.g., product design models). Despite most of the aspects being considered, the work \cite{42} ignores the critical issues such as the edge computing paradigm and advanced AI technologies.

In contrast, our AIaCM framework includes all the aspects in CM, including smart devices, smart interaction, AI technologies, and smart services. Meanwhile, our AIaCM framework also considers the advent of edge computing, software-defined networks, and advanced AI technologies. Moreover, we also present a full-fledged prototype to further demonstrate the effectiveness of the proposed framework (please refer to Section~\ref{sec:case} for more details). The implementation details of the AIaCM architecture are discussed next.

\section{Intelligent Manufacturing Devices}
\label{sec:devices}

\subsection{Edge computing-assisted intelligent agent construction}
In the customized production paradigm, manufacturing devices should be capable of rapid restructuring and reuse for small batches of personalized products~\cite{57,58}. However, it is challenging to achieve elastic and rapid control over the massive manufacturing devices. The agent-based system was considered a solution to this challenge~\cite{WShen:TSMC06,khan2019agent}. Agents can autonomously and continuously function in a collaborative system~\cite{59}. A multi-agent system can be constructed to take autonomous actions. Different types of agents have been constructed in~\cite{60,61,kovalenko2019model}.

\begin{figure}[!t]
  \centering
   \includegraphics[width=8.8cm]{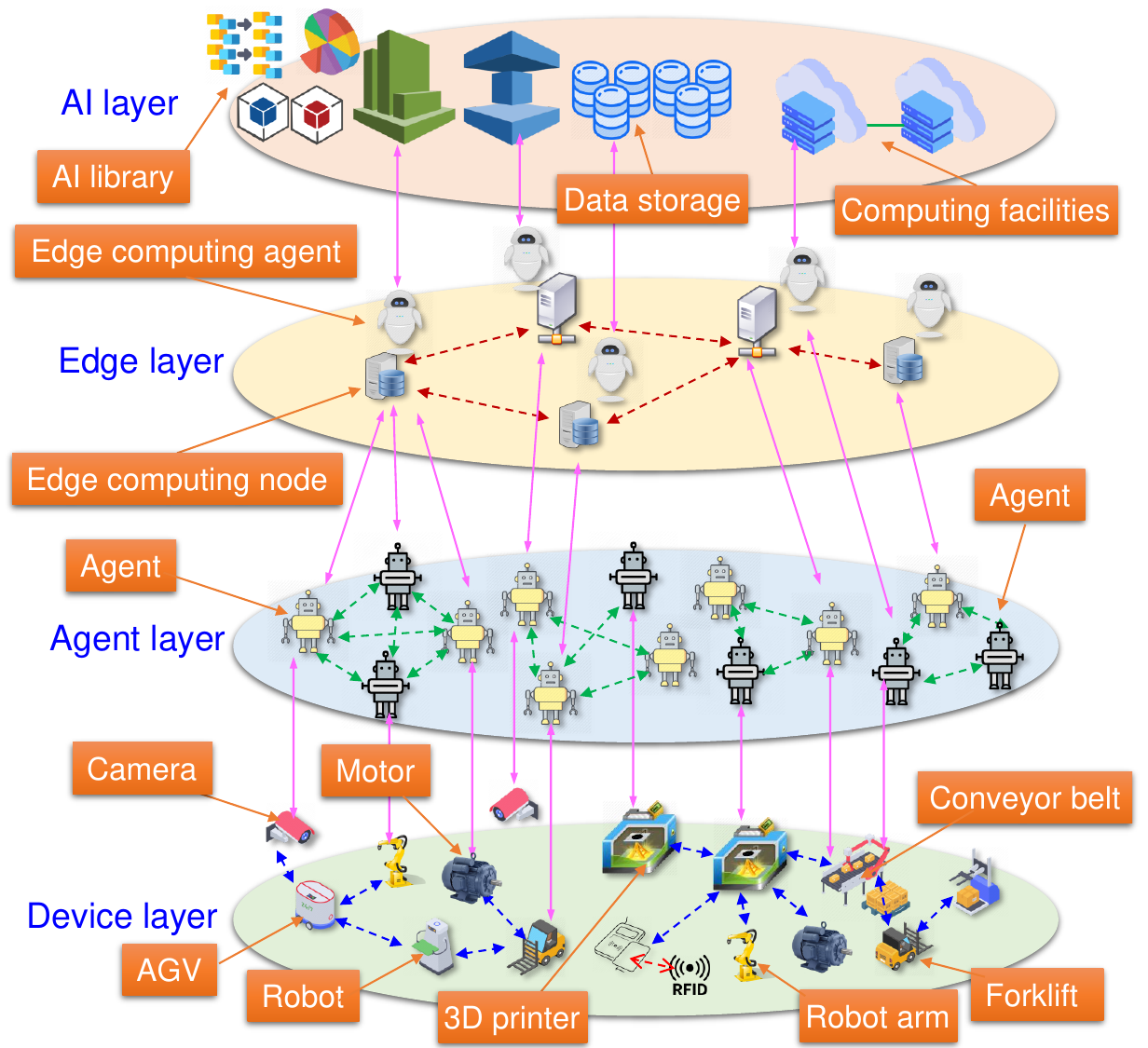}
        \caption{Edge computing-assisted manufacturing devices. This architecture includes the device layer, agent layer, edge computing layer, and AI layer.}
    \label{figure3}
\end{figure}

Although a single agent may have sensing, computing, and reasoning capabilities, it alone can only accomplish relatively simple tasks. Smart manufacturing may involve complex tasks, for instance, the image-based personalized product recognition, expected from the emerging multi-agent systems~\cite{62,63}. However, the multiple agents are deficient in processing massive data. Recent advances in edge computing can meet this emerging need~\cite{64,65,66}. As shown in Fig.~\ref{figure3}, a variety of decentralized manufacturing agents are connected to edge computing servers via high-speed industrial networks. The edge computing assisted manufacturing agents embrace the device layer, agent layer, edge computing layer, and AI layer.

An agent is equipped with a reasoning module and a knowledge base, offering basic AI functionalities such as inferencing and computing. Moreover, with the support of new communication technologies (e.g., 5G mobile networks and high-speed industrial wired networks), all agents and edge computing servers can be interconnected.

Agents run on edge computing servers to guarantee low-latency services for data analytics. The agent edge servers are connected by high-speed industrial IoT to achieve low latency. Generally, edge computing servers support a variety of AI applications.

An example of such a system is a personalized product identification based on deep learning image recognition. First, a multiple agent subsystem is constructed for producing personalized products. Then, a single agent records image or video data at different stages of the CM process. Next, the edge computing server runs the image recognition algorithms, such as a convolutional neural network (CNN), R-CNN, Fast R-CNN, Faster R-CNN, YOLO, or Single Shot Detection (SSD), all of which have demonstrated their advantages in computer vision tasks. The identification results are rapidly transferred to the devices. When the single edge computing server cannot meet the real-time requirements, the multiple agent edge servers may work collaboratively to complete the specific tasks such as product identification. Indeed, during the process, the master-slave or auction mode can be adopted for coordination, according to the status analysis of each edge server.

Additionally, with the help of edge computing, it is possible to establish a quantitative energy-aware model with a multi-agent system for load balancing, collaborative processing of complex tasks, and scheduling optimization in a smart factory~\cite{67}. The above procedure can also optimize the production line with better logistics while ensuring flexibility and manufacturing efficiency.

\subsection{Manufacturing resource description based on ontology}
Intelligent manufacturing will be greatly beneficial to the integration of distributed competitive resources (e.g., manpower and diverse automated technologies), so that resource sharing between enterprises and flexibility to respond to market changes are possible (i.e., CM). Therefore, in smart manufacturing, it is imperative to realize dynamic configurations of manufacturing resources~\cite{68,69}. CM can optimize lead time and manufacturing quality under various real-world constraints of dynamic nature (resource and manpower limitations, market demand, etc.). 

There are several strategies in describing manufacturing resources, such as databases, object-oriented method~\cite{zhang1999object}, and the unified manufacturing resource model~\cite{vichare2009unified}. In contrast to the conventional resource description methods, the ontology-based description is one of the most prominent methods. An ontology represents an explicit specification of a conceptual model~\cite{70}, by way of a classical symbolic AI reasoning method (i.e., an expert system). Modeling an application domain knowledge through an expert system provides a conceptual hierarchy that supports system integration and interoperability via an interpretable way~\cite{71,72}. 

\begin{figure}[!t]
  \centering
   \includegraphics[width=9.0cm]{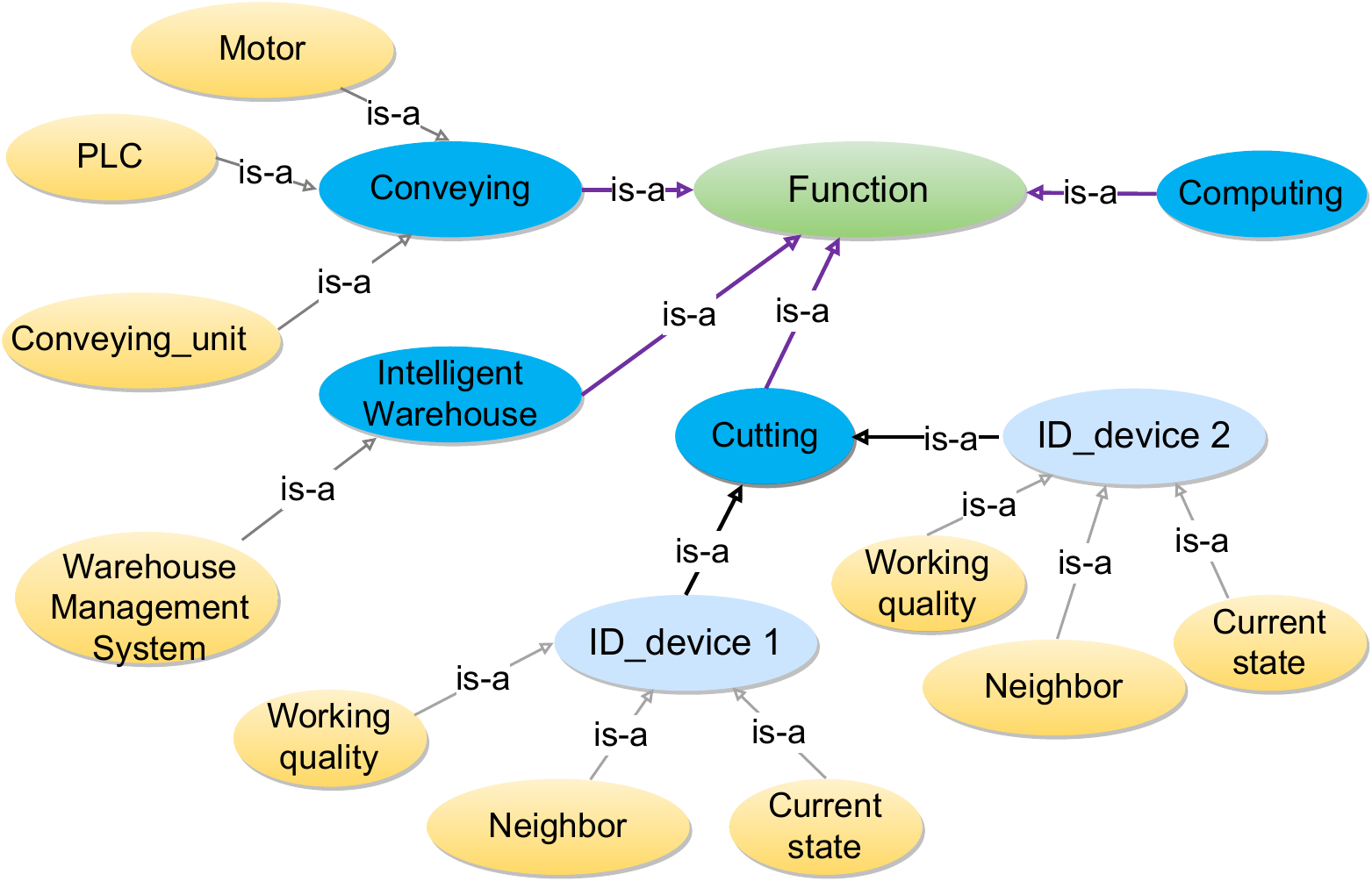}
        \caption{Manufacturing resources from the device function perspective. The CM resources of a product can be mapped into computing, cutting, conveying, and other functions.}
    \label{figure4}
\end{figure}
 In our previous work~\cite{73}, the device resources of smart manufacturing were integrated by the ontology-based integration framework, to describe the intelligent manufacturing resources. The architecture consisted of four layers, namely, the data layer, the rule layer, the knowledge layer, and the resource layer. The resource layer represented the entity of intelligent manufacturing equipment (e.g., manipulators, conveyor belt, PLC), which was essentially the field device. The knowledge layer was essentially the information model composed of intelligent devices, which was integrated into the domain knowledge base through the OWL language~\cite{74}. The rule layer was used to gather the intelligent characteristics of intelligent equipment, such as decision-making and reasoning. The data layer included a distributed database for real-time data storing, and the relational database was used to associate the real-time data.

Due to the massive amount of data generated from manufacturing devices, it is nearly impossible to consider all the manufacturing device resources. Thus, it is important to construct a new manufacturing description model to realize the reconfiguration of various manufacturing resources. In this model, the resources can be easily adjusted by running the model. Therefore, ontology modeling is conducted on a device and related attributes of an intelligent production line in CM. The manufacturing resources are mapped to different functions with different attributes. For instance, the time constraint of a product manufacturing is divided into a number of time slots with consideration of features of processes and devices. Then, the CM resources of a product can be mapped into computing, cutting, conveying, and other functions with the limited time slot, as shown in Fig.~\ref{figure4}. Next, a customized product can be produced by different devices with different time constraints. Accordingly, a product can be represented by ontology functions.

Meanwhile, after making a reasonable arrangement of different manufacturing functions at different time slots, a DL algorithm can forecast time slots of working states. The time slots of working states are important for the reconfiguration of manufacturing resources. Therefore, in actual applications, a different attribution of a device and customized products can be employed as a constraint condition.

\subsection{Edge Computing in Intelligent Sensing}
\label{subsec:edge-sensing}
The concept of ubiquitous intelligent sensing is a cornerstone of smart manufacturing in the Industry 4.0 framework. Numerous research studies have been conducted in monitoring manufacturing environments~\cite{75,76,77}. Most published results adopt a precondition-sensing system that only accepts a static sensing parameter. Obviously, this results in inflexibility and the sensing parameters are difficult to be adjusted to fulfill different requirements. Second, although some studies claim dynamic parameter tuning, the absence of a prediction function is still an issue. Existing environment sensing (monitoring) cannot adjust the sensing parameters in advance to achieve a more intelligent manufacturing response.

\begin{figure}[!t]
  \centering
   \includegraphics[width=8.8cm]{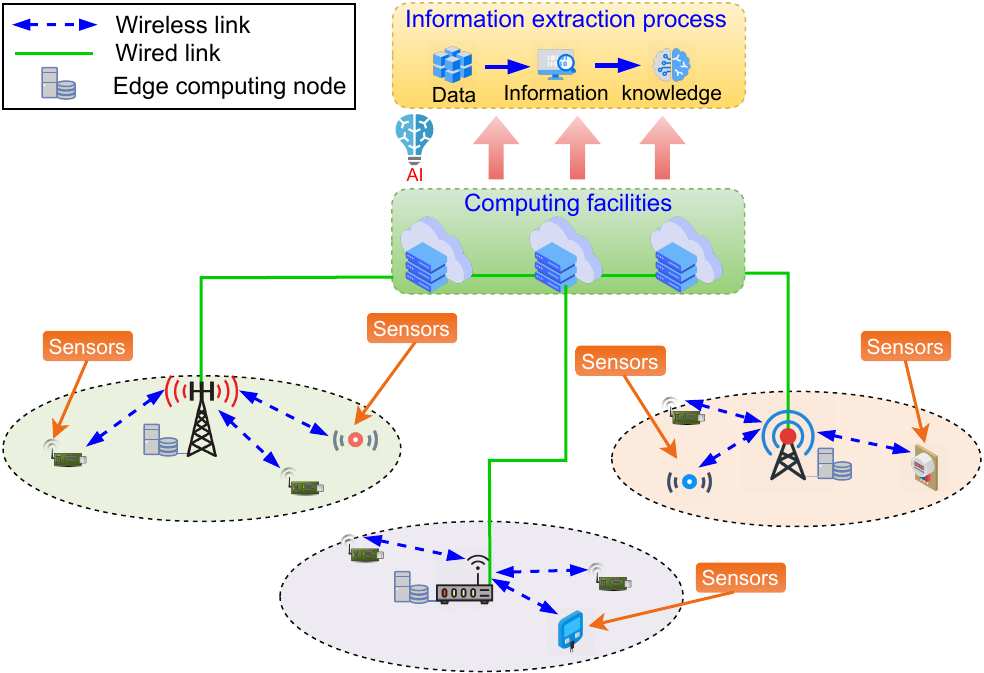}
        \caption{Intelligent sensing based on edge AI computing. Sensor nodes collect ambient data while edge computing notes can preprocess and cache the collected data, which can be further transferred to remote cloud servers for in-depth data analysis.}
    \label{figure5}
\end{figure}
As shown in Fig.~\ref{figure5}, the manufacturing environment intelligent sensing based on the edge AI computing framework includes two components: sensors nodes and edge computing nodes~\cite{78}. Generally, smart sensor nodes are equipped with different sensors, processors, and storage and communication modules. The sensors are responsible for converting the physical status of the manufacturing environment into digital signals, and the communication module delivers the sensing data to the edge server or remote data centers. The edge computing servers (nodes) include the stronger processing units, larger memories, and storage space. These servers are connected to different sensors nodes and deployed in approximation to the devices, with the provision of the data storage and smart computing services by running different AI algorithms. Meanwhile, the edge computing servers are interconnected with each other to exchange information and knowledge.

Especially, the sensing parameters can be adjusted in a flexible monitoring subsystem in the manufacturing environment, according to different application requirements and the task priority. To achieve a rapid response high priority system, the edge AI servers should have access to the sensing data, and capability to categorize the status of the CM environment. This can be done by processing the data features through ML classification algorithms such as logistic regression, SVM, and classification trees. When the data is out of the safety range, a certain risk may exist in the manufacturing environment. For instance, if an anomalous temperature event would happen in the CM area, the edge server could drive the affected nodes to increase their temperature sensing frequency, in order to obtain more environmental details and to make proactive forecasts and decisions.

The environmental sensing data delivery is another important component in CM. With the development of smart manufacturing, a sensing node not only performs sensing but also transmits the data. With the proliferation of massive sensing data, sensor nodes have been facing more challenges from the perspectives of data volume and data heterogeneity. In order to collect environment data effectively, it is needed to introduce new AI technologies. The sensor nodes can realize intelligent routing and communications by adjusting the network parameters, assigning different network loads and priorities to different types of data packets. With this optimized sensing transfer strategy, the AI methods can make adequate forecasts with reduced bandwidth usage.

\textbf{Discussion.} We present intelligent manufacturing devices from edge computing-assisted intelligent agent construction, manufacturing resource description based on ontology, and edge computing in intelligent sensing. It is a challenge to upgrade the existing manufacturing devices to improve the interoperability and the inter-connectivity. Retrofitting instead of replacing all the legacy machines may be an alternative strategy in this regard. The legacy manufacturing equipment can be connected to the Internet by additively mounting sensors or IoT nodes in approximation to existing manufacturing devices~\cite{KZhang:TII20,JHuang:TII20}. Moreover, monitors can be attached to existing machinery to visualize the monitoring process. It is worth mentioning that retrofitting strategies may apply for the sensing or monitoring scenarios while they are not suitable or less suitable for the cases requiring to make active actions (like control or movement). Furthermore, a comprehensive plan should be made in advance rather than arbitrarily adding sensors to the existing production line~\cite{pueo2019design}. Retrofitting strategies also have their limitations, such as a limited number of internal physical quantities can be monitored in a retrofitted asset with respect to a newly-designed smart machine. 

\section{Intelligent Information Interaction In a Smart Factory}
\label{sec:interaction}
In the CM domain, the information exchange system needs to fulfill the dynamic adjustment of network resources so as to produce multiple customized products in parallel. In order to obtain optimal strategies, many studies have focused on this topic, and proposed insightful algorithms as well as strategies~\cite{79}. However, there are still two open issues: a network framework to dynamically adjust network resources, and the end-to-end (E2E) data delivery. In this section, we present software-defined industrial networks and AI-assisted E2E communication to tackle these two challenges.

\begin{figure}[!t]
  \centering
   \includegraphics[scale=0.6]{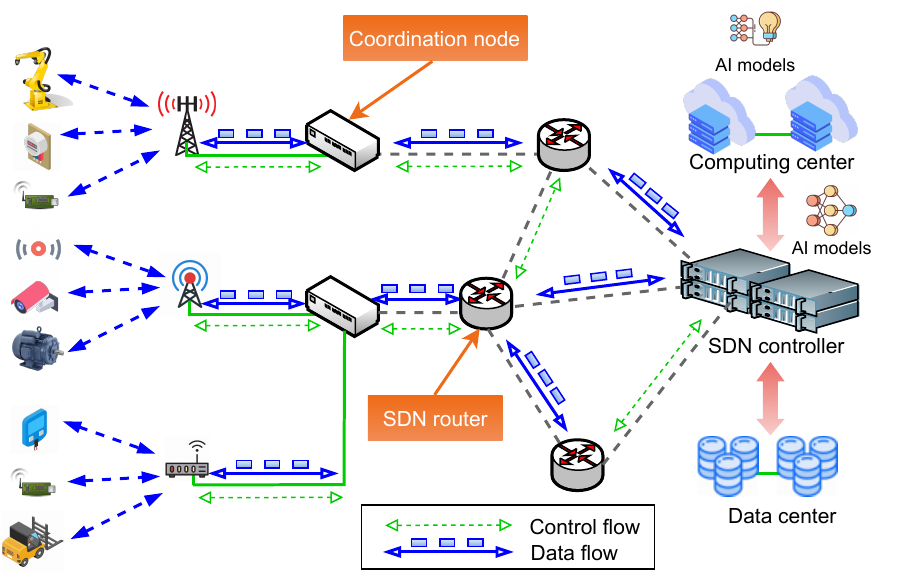}
        \caption{Software-defined industrial networks consist of network coordinated nodes, SDN routers, SDN controllers, data centers, and cloud computing servers which can support intensive computing tasks of AI algorithms.}
    \label{figure6}
\end{figure}

\subsection{Software-defined industrial networks}
Industrial networks are a crucial component in CM, and customized product manufacturing groups can be understood as subnets. Via an industrial network (consisting of base stations, access points, network gateways, network switches, network routers, and terminals), the CM equipment and devices are closely interconnected with each other and can be supported by edge or cloud computing paradigms~\cite{80}. Taking full advantage of AI-driven software-defined industrial networks, and relevant networking technologies is an important method to achieve intelligent information sharing in CM~\cite{81,82}.

In conventional industrial networks, network control functions have been fixed at network nodes (e.g., gateways, routers, switches). Consequently, industrial networks cannot be adapted to dynamic and elastic network environments, especially in customized manufacturing. The software-defined networking (SDN) technology can separate the conventional network into the data plane and the control plane~\cite{9732420}. In this manner, SDN can achieve flexible and efficient network control for industrial networks. It has been reported that a software-defined industrial network can increase the flexibility of a dynamical network system while decreasing the cost of constructing a new network infrastructure~\cite{83}. 

The introduction of AI technologies to SDN can further bestow network nodes with intelligence. As demonstrated in Fig.~\ref{figure6}, AI technologies are introduced into traditional SDN so as to form a novel software-defined industrial network (SDIN). The proposed SDIN contains a number of mapping network nodes, SDIN related devices, data centers, and cloud computing servers to support intensive computing tasks of AI algorithms. Manufacturing devices are connected by their communication modules, and they are mapped to different network terminal nodes. On the SDIN level, key devices such as coordinated nodes and SDIN controllers construct the SDIN layer. First of all, coordinated nodes are linked with the ordinary nodes, and deliver network control messages from other SDN devices. Second, the SDN routers are the key devices that realize the separation of data flow and control flow of the entire manufacturing network. In addition, the SDIN controller is directly connected to the AI server, and the AI server provides network decisions directly to the SDN controller.

In the network information process, AI algorithms, such as deep neural networks, reinforcement learning, SVM, and other ML algorithms can be executed in a server according to the state of the network devices, such as load information, communication rate, received signal strength indicator, and other data. Then, the AI server returns the optimized results to the SDN controller, and the results are divided into different instructions for different network devices in the light of a specific CM task. Following the above steps, the SDN controllers send a set of instructions to the routers and the coordination nodes. Finally, network terminals readjust the related parameters, (e.g., communication bandwidth, transmitted powers) to complete the data communication process.

Intelligent optimization algorithms (e.g., ant colony or particle swarm optimization) can find optimal data transfer strategies – based on the network parameters provided by the SDIN, or given by the constraints of data interaction. These algorithms can adjust the latency and energy consumption requirements. Thus SDIN can improve the information management processes within a CM industry framework, reducing the cost of dynamically adjusting or reconfiguring network resources. Moreover, it can improve and propel the whole manufacturing intelligence. Additionally, by adopting an AI-assisted SDIN, the production efficiency can be further improved.

\subsection{End-to-End communication}
End-to-end (E2E) or device-to-device communication between manufacturing entities is a convenient communication strategy in industrial networks~\cite{84,85}. E2E communication provides communication services with lower latency and higher reliability, as compared to a centralized approach~\cite{86}. With effective information interaction via E2E communication, the entire system can achieve full connectivity. In the context of CM, data transmission with different real-time constraints has become a critical requirement~\cite{87}. The E2E industrial communication approach optimizes the usage of network resources (e.g., network access and bandwidth allocation) through data communication of varying latency~\cite{88,89}. Meanwhile, in order to realize the E2E communication in the industrial domain, a hybrid E2E communication network – based on the AI technology and SDIN – is here constructed by exploiting different media, communication protocols, and strategies. The hybrid E2E-based communication mechanism with the AI assistance can be divided into three layers: the physical layer, the media access controlling (MAC) layer, and the routing layer.

\begin{figure*}[!t]
  \centering
   \includegraphics[width=17.3cm]{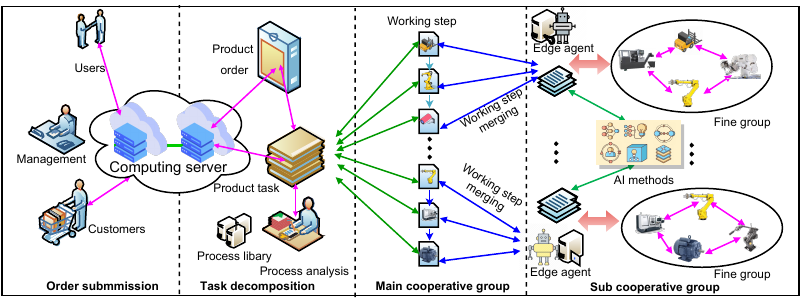}
        \caption{Cooperative multiple agents. The strategy of cooperative multiple agents can be divided into i) the order of submission, ii) task decomposition, iii) cooperative group and iv) subgroup assignment.}
    \label{figure7}
\end{figure*}

In the physical layer, according to the advantages and disadvantages of the involved communication technologies, different communication media include optical fiber~\cite{90}, network cable~\cite{91}, and wireless radio~\cite{92}. Generally, industrial communications can be divided into wired or wireless communications. On the one hand, wired communications typically exhibit high-stability and low-latency. A representative case is an industrial Ethernet, which is based on a common Ethernet and runs improved Ethernet protocols, such as EtherCAT~\cite{93}, EtherNet/IP~\cite{94}, and Powerlink~\cite{95}. On the other hand, wireless networks have been adopted in applications with relatively high flexibility~\cite{96,97}. Nowadays, an increasing number of mobile elements have been incorporated in manufacturing systems; therefore, wireless media has been widely exploited in mobile communications~\cite{98}. Conventional strategies on fixed and static industrial networks may not fulfill the emerging requirements on flexible network configurations. The AI and related technologies, such as deep reinforcement learning, optimization theory and game theory, can play significant roles in improving the communication efficiency in the physical layer, e.g., determining the optimal communication between wired and wireless networks while achieving a good balance between network operational cost and network performance.

In the MAC layer, different devices have different requirements for E2E communications according to their specific functions. Although many different MAC protocols have been proposed (e.g., CSMA–Carrier Sense Multiple Access)~\cite{99}, CDMA–Code Division Multiple Access)~\cite{100}, TDMA–Time Division Multiple Access)~\cite{101} and their improved versions, these methods still lack flexibility, and do not fulfill the emerging requirements of industrial applications. Generally, industrial E2E communications can be divided into two categories: periodic communications and aperiodic communications. Similarly, AI plays an important role in the MAC layer. An example is a hybrid approach that combines the CSMA and TDMA, with an intelligent optimization method, to improve the efficiency of the E2E communication. In particular, the two categories of communication requirements (high and low real-time or periodic and aperiodic communications) are classified by the AI-based method (e.g., naïve Bayes). Next, an improved hybrid MAC is constructed on top of the CSMA and TDMA. TDMA and CSMA schemes deal with the periodic and aperiodic data flows of the E2E communications. The size of this proposed mechanism can be adjusted in accordance with the AI-optimized results of a real application.

The network routing is also another key component of E2E communications. The key node of the routing path plays an important role in the E2E communications as well. However, the performances of routing key nodes are impacted by the workload; for instance, the amount of forwarded data. Similarly, AI plays a significant role in the routing layer. The predicted state parameters, such as communication rate and network loads of key nodes, can be obtained by using historical data from the network node status by algorithms, such as deep neural networks or deep reinforcement learning (e.g. deep Q-learning). 

\section{Flexible Manufacturing Line}
\label{sec:manu-line}
A flexible manufacturing production line realizes customization. AI-driven production line strategies and technologies, such as collective intelligence, autonomous intelligence, and cross-media reasoning intelligence, have accelerated the global manufacturing process. Therefore, the subjects of cooperative operation between multiple agents, dynamic reconfiguration of manufacturing, and self-organizing scheduling based on production tasks are presented in this section.

\subsection{Cooperative multiple agents}
Cooperation among multiple agents is necessary to dynamically construct collaborative groups for the completion of customized production tasks~\cite{102}. As discussed in Section~\ref{sec:devices}, multiple agents with edge computing provide a better option than a single device to build a collaborative operation to realize CM~\cite{67,103}. Therefore, by combining the edge computing-assisted intelligent agents and different AI algorithms, a novel cooperative operation can be constructed as shown in Fig.~\ref{figure7}. The strategy of cooperative operation by multiples agents can be divided into the order of submission, task decomposition, cooperative group, and subgroup assignment.

The working process of a flexible manufacturing production line can be described as follows. First, according to the customers' requirements, the CM product orders are issued to the manufacturing system through the recommender system. After receiving the product orders, the AI-assisted task decomposition algorithms take the product orders as the input, the device working procedure as the output, and the product manufacturing time as a constraint; these algorithms are mainly executed at the remote cloud server. A product order can be divided into multiple subtasks, which are sent to all the agents via the industrial network. After the negotiation, agents return the answers to the edge server, which handles the working subtasks according to corresponding conditions and constraints. Next, the AI-assisted cost-evaluation algorithm calculates the cost of a producing group (i.e., cooperative manufacturing group) from the historical data. Then, the edge agents intelligently select suitable device agents to finish the product order after considering the whole cooperative group performances, such as producing time and product quality. Moreover, the edge agents send the selection result to the device agents, which are chosen to take part in the producing order. The main cooperative group is constructed based on the working steps.

The main cooperative group may not be well suited for real applications, especially for complicated CM tasks. Therefore, an AI-based method for constructing a suitable-size cooperative subgroup is an important step for dealing with the mentioned problem. A possible strategy is to use cognitive approaches such as the Adaptive Control of Thought—Rational (ACT-R) model~\cite{taatgen_lebiere_anderson_2005}. These subtasks cooperative groups can be mapped to the digital space (i.e., edge agent) and form even lower level subgroups, all interconnected by the conveyor, logistics systems, and industrial communication systems. Each subgroup can delegate the same edge agent, to provide the management and customers with manufacturing services. The characteristics of the subgroups are partly derived from the process constraints and the physical constraints of the plant. In principle, the higher the constrains the deeper the task tree will expand, from more abstract tasks to particular atomic targets achievable by the present devices. This structure can be replicated with a probabilistic graphical model or with a fuzzy tree.

After all the agents have been assigned with subtasks, they form two level-cooperative groups. The formation of these cooperative groups is beneficial to resource management. Then, according to the manufacturing task attributes, multiple agents complete the producing task. During this period, the corresponding device agents send their status data to edge servers timely, and the manufacturing process can be monitored by analyzing these data in the entire system. In contrast to the AI-driven cooperative operation between multiple agents, conventional methods often rely on human operators who participate in the whole process or computer-assisted operators also requiring human interventions. These methods inevitably result in huge operational expenditure. 

\subsection{Dynamic reconfiguration of manufacturing Systems}
With the scientific development of the industrial market and manufacturing equipment, different industrial devices present different performance requirements representing multiple function trends~\cite{104}. For instance, the latest Computer Numerical Control (CNC)  machine tool can complete a wide range of tasks, from lathing to milling functions. On the other hand, a dedicated manufacturing line does not meet new industrial requirements, especially for customized production~\cite{105}. The trend today is towards reconfiguration and reprogrammability of manufacturing processes~\cite{106}. Although several studies have investigated the problem and presented meaningful results~\cite{107,108}, most of them lack intelligent design to fulfill the emerging requirements of dynamic reconfiguration of manufacturing systems, especially for customized manufacturing. In particular, the work~\cite{107} focuses on the communications between agents while \cite{108} investigates the relationship between manufacturing flexibility and demands. Thus, AI technologies have seldom been adopted in these studies. At present, ontology (as shown in Section~\ref{sec:devices}) offers insights into dynamic reconfiguration of manufacturing resources~\cite{86,109}.

\begin{figure}[!t]
  \centering
   \includegraphics[width=8.8cm]{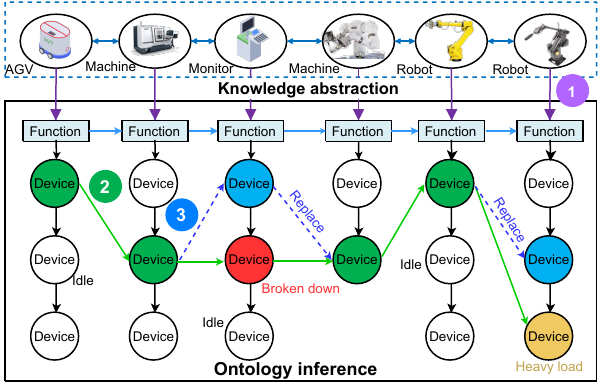}
        \caption{Dynamic reconfiguration of manufacturing resources: 1) device selection by ontology reasoning based on the device function; 2) CM production line is constructed; 3) automatic switching to other devices from heavy-load devices or broken devices.}
    \label{figure8}
\end{figure}

A schematic of the dynamic reconfiguration process based on the ontology inference is shown in Fig.~\ref{figure8}. Each customized product invokes several processing procedures. First, a personalized product manufacturing-related device (such as cutting, materials handling device) is selected by ontology reasoning based on the device function. Then, the second selection of the devices involved in the manufacturing is finished according to ontology results with respect to the related manufacturing process, the manufacturing time, manufacturing quality, and other parameters of a device. Finally, a CM production line is constructed. Specifically, when the production line receives a production task, the raw material for a specific type of products is delivered from an autonomous warehouse. Then, the production line completes the manufacturing tasks in the process sequence. Furthermore, when one of the manufacturing devices breaks down during the process, automatic switching of the related machining equipment by ontology inference is conducted. Meanwhile, the reasoning mechanism reflects the reconstruction function of a flexible production of the production line.

The presented approach leads to optimal process planning and functional reconstruction. Besides, it shows the strengths of ontology modeling and reasoning. In practice, only ontology and constraints need to be established according to the above description. According to Jena syntax\footnote{Jena syntax defines a set of rules, principles, and procedures to specify the semantic web framework of Apache Jena (\url{https://jena.apache.org/getting_started/index.html}).}, the corresponding API interface can be invoked to meet the task requirements of this scenario. In the future, other AI algorithms are expected to be integrated with ontology inference. 

\begin{figure}[!t]
  \centering
   \includegraphics[width=8.9cm]{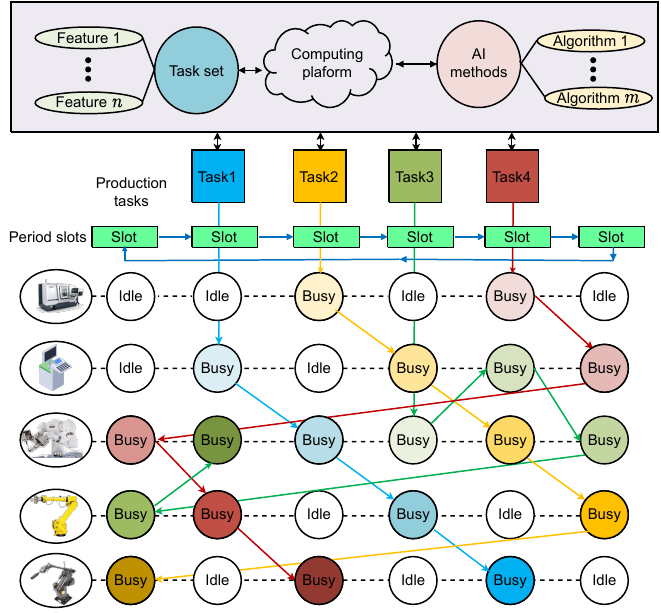}
        \caption{Self-organization of schedules of multiple production tasks consists of three steps: task analysis, task decomposition and task execution.}
    \label{figure9}
\end{figure}

\subsection{Self-organizing Schedules of Multiple Production Tasks}
Product orders generally have stochastic and intermittent characteristics as the arrival time of orders is usually uncertain~\cite{110}. This may result in having to share production resources among multiple tasks. Therefore, creating self-organizing schedules with a time slot based on multiple agents for multiple production tasks is paramount~\cite{61}. The mechanism of self-organizing schedules for multiple production tasks can be divided into three steps: task analysis, task decomposition, and task execution.

As shown in Fig.~\ref{figure9}, in terms of initialization, when a new production task is processed by the multi-tasking production line, the new production tasks are divided into multiple steps by an AI-based method executed at the cloud. Additionally, according to the process lead time, the producing period can be decomposed into time slots of different lengths. Moreover, for one working step in a time slot, edge agents select all idle device agents by comparing the mapping relationship between working steps and device agent functions. This processed time slot information is then broadcasted to all the agents simultaneously. Then, idle device agents choose the working step by price bidding or negotiating with others, according to the manufacturing requirements and self-conditions (e.g., manufacturing time and quality). These results are broadcasted to other agents, including different servers. Next, the edge agents update the working state of the idle device agent in the corresponding time slot. These procedures are repeated until the new task steps are allocated within a certain or fixed time. Lastly, multiple agents finish the scheduling of the new production task in a self-organization manner.

\begin{figure*}[!t]
\centering
\subfloat[Framework for case study]{\includegraphics[width=3.4in]{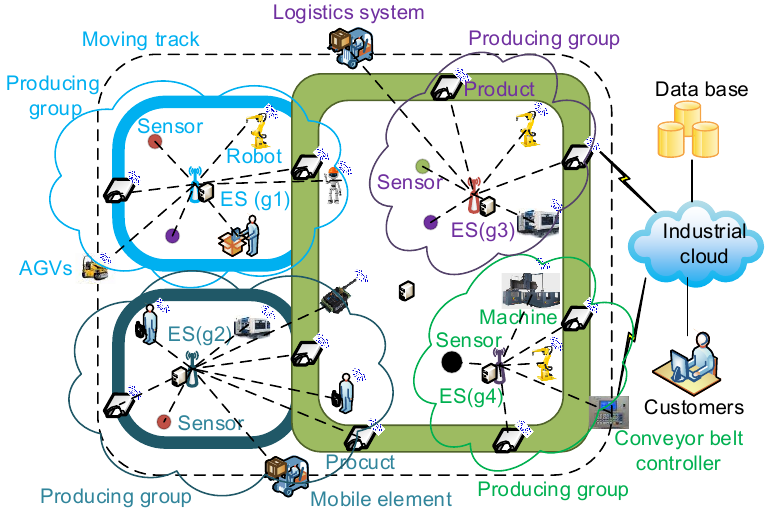}%
\label{fig_first_case}}
\hfil
\centering
\subfloat[Implementation scenario of CM]{\includegraphics[width=3.6in]{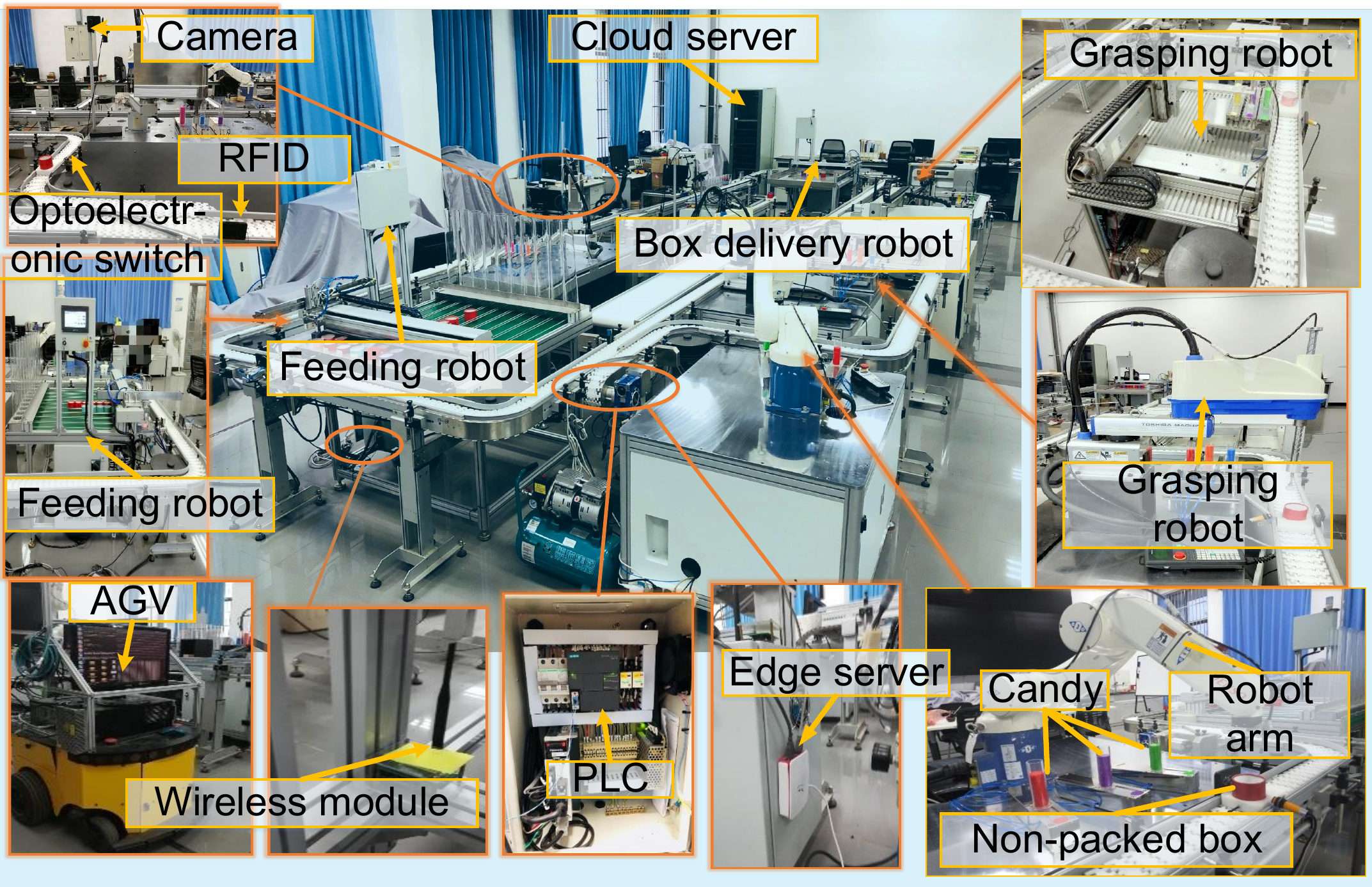}%
\label{fig_second_case}}
\caption{The framework of customized candy wrapping smart production. (a) The framework includes CM devices, the industrial network, a conveyor, and a cyber-physical system. (b) The implementation of customized candy wrapping line consists of diverse devices.}
\label{fig_sim}
\end{figure*}

Self-organization of schedules with multiple agents and time slots can effectively complete simultaneous production tasks using a flexible production. Furthermore, production line efficiency is improved. Consequently, all manufacturing resources, including different devices and subsystems, are more intelligent to finish the multiple production tasks autonomously. In contrast, conventional methods often require huge human resources in scheduling and planning production tasks~\cite{MCKAY20035}. Despite the recent advances in computer-aid methods~\cite{BIEL2016243}, they still require substantial human interventions and cannot meet the flexible requirements. 

However, we have to admit that AI-driven self-organization of schedules does not get rid of humans in the loop of the entire production process. The main goal of AI-driven methods is to save unnecessary human resource consumption and mitigate other operational expenditures. In this manner, human workers can concentrate on planning and optimizing the overall production procedure instead of conducting tedious and repetitive tasks. Meanwhile, an appropriate human intervention is still necessary when full automation is not achievable or is partially implemented. In this sense, AI-driven methods can also assist human workers to give intelligent determinations.

\section{Case Study}
\label{sec:case}
In the section, a case study is presented, showcasing the following aspects: prototype platform construction, big data analysis using AI technology for preventive maintenance, and cloud-assisted customization service.

\subsection{Prototype platform construction}
We implement a prototype of the AI-assisted CM framework, namely a customized candy wrapping production line. As shown in Fig.~\ref{fig_first_case}, the framework includes the following components: CM devices, the industrial network, a conveyor, and a cyber-physical system. All components are connected through the industrial network, i.e., OPC Unified Architecture (OPC UA) and Data Distribution Service (DDS). Fig.~\ref{fig_second_case} illustrates the implementation of customized candy wrapping line. The candy packing line mainly includes the production stations and the logistics transmission system. In the logistics transmission system, the packing box is continuously transferred by the conveyor belts or AGVs. The production stations are distributed discretely between the mainline and the branch line, and RFID tags are adopted to obtain the operation information. The equipment types of the production stations include the materiel feeding, candy grasping, box delivery, and finished goods storage. The presented system meets the requirements of small-batch production. In particular, the packaged candy followed the taste, flavor and color preferences given by the customers. The system includes four layers, all of which are connected by the industrial IoT with different link functions.

The first layer is the device layer, including five robots, two AGVs, a conveyor system, and a warehouse. The device layer performs the basic functions of an intelligent production line, such as carrying, clipping, loading raw material, and unloading final products. Cognitive robots can be vertically integrated into a cyber-physical system in smart manufacturing~\cite{111}.

The industrial network layer (second layer) plays a key role in the information interaction and intelligent connection of different communication technologies – e.g., industrial wireless local networks (Wi-Fi, ZigBee), industrial Ethernet, industrial NFC (Near Field Communication), and mobile communications. There are three sub-networks for finishing different latency communication functions~\cite{112}. Specifically, wired industrial networks are employed as the inner equipment to achieve higher real-time performance. In this aspect, the wireless industrial networks were mainly adopted in the monitoring system while the mobile wireless local networks also helped to achieve higher-level flexibility~\cite{113}; for instance, mobile wireless nodes were dynamically deployed to monitor the industrial environment status.

The third layer is the computing layer, which is mainly involved with the analysis, computing, and knowledge mining of big data. A commercial solution has been adapted to build a cloud platform. XenServer developed by Citrix is used to realize the virtualization of the server cluster consisting of multiple virtual machines and the management of virtual machines. Meanwhile, we also establish a big data analytics framework, which is a software architecture based on a cloud platform for big data storage and distributed computing.  Apache Hadoop, an open-source solution, is used to provide the non-relational database HBase and the computing architecture of YARN (Yet Another Resource Negotiator). On top of the big data framework, the AI-assisted optimization algorithms (such as deep learning models) have been deployed to realize intelligent applications. To meet different latency requirements in the platform, a hybrid computing paradigm, orchestrating the cloud and edge computing paradigms, is adopted. Explicitly, edge computing is used to deal with real-time tasks, while cloud computing was focused on completing time-insensitive tasks, such as historical data processing. The deployment of edge computing enables cloud service characteristics such as mobile computing, scalability, and privacy policy~\cite{114}.

The fourth layer is the service layer. In this level, a large number of manufacturing resources are stored at the cloud platform, which offers different AI services. Pattern recognition, accurate modeling, knowledge discovery, reasoning, and decision-making capabilities are provided.

\begin{figure*}[!t]
  \centering
   \includegraphics[width=12.6cm]{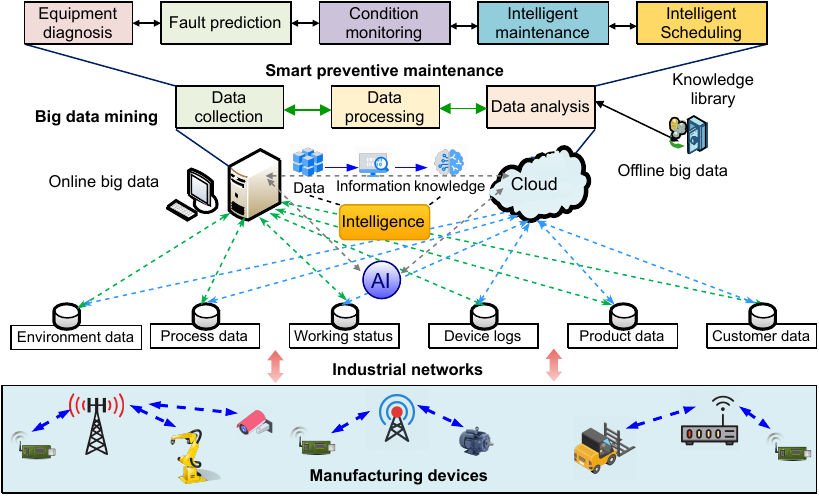}
        \caption{Big data analysis using AI technology for smart preventive maintenance. The smart preventive maintenance system consists of CM devices, industrial networks, big data processing center, and applications for preventive maintenance. The data pipeline consisted of three main steps, data collection, data processing, and data mining and analysis using AI. }
    \label{figure11}
\end{figure*}

The working process of the platform is as follows. First, customers select candy products according to their preferences, which included the color, taste, quantity, and variety of the candies in an AI recommender web service system. Then, these proposed schemes and candy order parameters are delivered to the manufacturing cloud through the web service, and the web server was connected to the cloud via the Internet. The related product orders are created according to the submitting information. These orders were decomposed into different working steps by the ontology-based manufacturing system. Next, the multiple agents completed the production tasks in a self-organized way. After obtaining the working steps, the manufacturing devices are assembled into collaborative groups to finish all tasks. Thereafter, the platform finished the candy wrapping task.

During the product manufacturing process, the manufacturing data is collected by sensors and is then transmitted to the cloud or nearby edge servers. The analyzed results provide key information for product monitoring. More importantly, these results can be used to adjust the processes and procedures to ensure higher quality and increase the production efficiency of the whole system. The model-driven method with ontology proposed in~\cite{115} was used to achieve interoperability and knowledge sharing in a manufacturing system across multiple platforms in the product lifecycle. When multiple tasks were needed to be finished in the platform, the manufacturing resource reconstruction methods were employed for production scheduling. The cloud-based manufacturing semantic model proposed in~\cite{116} was used to obtain general task construction and task matching. After implementation, three candy-wrapping tasks with ten different candies were processed in the AI-assisted platform at the same time, which represented a typical production line model for mass wrapping, and the first-in-first-out (FIFO) scheme was adopted accordingly.

\subsection{Big data analysis using AI technology for preventive maintenance}
Preventive maintenance for smart manufacturing has received attention in the literature~\cite{117,118,119,120,121,122}. A system architecture for an active preventive maintenance system was proposed in~\cite{123}. Based on this architecture, an improved preventive maintenance mechanism was constructed by merging cloud computing and edge computing with deep learning, as shown in Fig.~\ref{figure11}. The smart preventive maintenance system was composed of CM devices, industrial networks, big data processing centers, and applications for preventive maintenance. The data pipeline consisted of three main steps, data collection, data processing, and data mining and analysis using AI.

From the perspective of preventive maintenance, the related data collection represented the fundamental step in the following analysis. Different data, including the environmental data, product processing data, device working status, and device logs were collected and transmitted to the computing servers, such as cloud and edge computing servers, through the industrial network or industrial IoT. Edge computing and cloud computing paradigms presented in~\cite{124} were employed to address elastic and virtual manufacturing resources, which provided opportunities for real-time monitoring of production Key Performance Indicators (KPIs) and smart inventory management. The computing servers were responsible for data processing and for device maintenance. The data includes different types of manufacturing device-related data. First, redundant and misleading data were removed during the data collection process. Then, the abstract data for real-time or historical big data analysis was used for equipment maintenance. The AI-based techniques (e.g., deep learning) have been regarded as the most effective way for system and equipment fault recognition based on big data analysis, so these techniques were adopted in this step. Consequently, the manufacturing maintenance knowledge base was built on top of big data. Note that AI techniques play an important role in extracting knowledge from massive and heterogeneous manufacturing data, consequently achieving the intelligence~\cite{Fricke:2018}. The manufacturing data includes 1) structured data such as data stored in a rational database and 2) non-structured data such as text, documents, sound, image, and video. Diverse AI algorithms such as ontology learning, natural language processing, and deep learning can be leveraged.

Furthermore, in our previous work~\cite{123}, a big data solution for active preventive maintenance in manufacturing environments was proposed. This approach essentially combined a real-time active maintenance mechanism with an offline prediction method. The real-time performance was considered as the main feature of a manufacturing system, especially equipment maintenance. Therefore, to achieve the system maintenance tasks, the hybrid edge and cloud computing paradigm for big data analysis represented a better option for preventive maintenance. Specifically, the equipment maintenance tasks can be divided into online and offline tasks. On the one hand, as edge servers being deployed close to the equipment can provide low-latency (real-time) service, edge computing was adopted for dealing with the data online. On the other hand, cloud servers have powerful computing capabilities, and offline big data processing was executed at the cloud layer. Moreover, the analysis result was delivered to the management or data visualization system. The segmented model provided in~\cite{125} for preventive maintenance of semiconductor manufacturing equipment, including both parametric and non-parametric models, was used for preventive maintenance.

Consequently, big data with AI is a key technology for equipment maintenance in smart manufacturing. Big data helps to build comprehensive condition monitoring and prediction systems, which can provide preventive maintenance scheduling according to the equipment status. Integrated with the AI-based methods, big data analysis can construct a maintenance knowledge library and decrease the cost of operation and maintenance management of a smart manufacturing system.

\subsection{Cloud-assisted customization service}
Unlike traditional manufacturing, CM can provide customized services. In other words, customers can participate in the process of intelligent manufacturing~\cite{126,127}. Recently, cloud computing has been proven to provide support for customers taking part in the production process and drive the customization services in a seamless manner~\cite{112,128}, allowing different data services to be quickly accessed by customers.

Therefore, the integration of cloud computing with customization services can improve the user experience of customization services. We name such integration of cloud computing with customization services as cloud-assisted customization services.
As shown in recent studies~\cite{129,ZHANG201912}, cloud-assisted customization services are well suited to offer a better user experience to the customer. In particular, cloud-assisted customization services are user-centric, demand-driven, and service-oriented. 

\begin{figure}[!t]
  \centering
   \includegraphics[width=8.3cm]{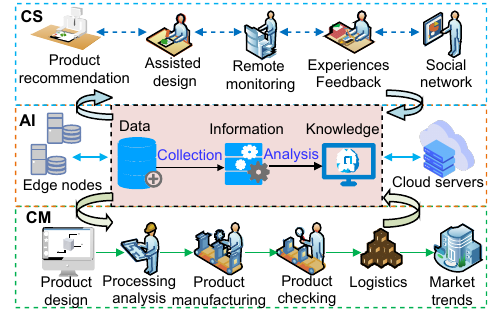}
        \caption{The cloud-assisted customization services cover the entire produce life cycle: early stage, middle stage and later stage.}
    \label{figure12}
\end{figure}

Based on the multiple different services provided by cloud computing, the customization service can be achieved from the production cycle perspective, as shown in Fig.~\ref{figure12}. The entire production cycle was divided into three stages: early stage, middle stage, and later stage. In the early stage of product production, customers can select more suitable products by the intelligent recommender system. As for the intelligent recommender system, big data analysis is adapted to integrate the order data, the production data, and the packing line status. Spark MLlib is adapted to realize the personalized recommendation, which has advantages on algorithm, experience, and performance.  Moreover, using cloud-assisted design, customers can also design the manufacturing products without a professional background in a digital simulation environment, virtual reality (VR), and/or augmented reality (AR). In the middle stage of the production, CM can provide other personal customization services. For instance, customers can remotely monitor the details of the production process via a digital twin or a virtual production line system. Then, logistics information of different stages can be sent to specific users via cloud computing and mobile Internet. At the later stage, customers can provide feedback on their user experience; this feedback can be used to improve the manufacturing process.

An important difference between intelligent manufacturing and traditional manufacturing lies in the fact that users can participate in the production process via cloud-assisted technologies in intelligent manufacturing. Through industrial cloud computing, CM can provide customers and users with customization services. This type of customization service is to motivate participation and construct a new production shopping experience.

In terms of the market trend prediction, the more the market data gathered, the more accurate the results. Different algorithms can assist in the data-driven decision-making process. For example, shape mining is a framework based on engineering design data~\cite{130} and has been applied to passenger car design. Further, the apriori and C5.0 algorithms for data mining are able to extract rules from databases~\cite{131}.

The platform used in the experiment, cannot only finish on-demand candy production based on the customization parameters but also adapt to changes in the market. The constructed platform can increase the efficiency of the entire system of candy wrapping. This methodology, in combination with the aforementioned decision-making algorithms, constitutes a proof of concept of a seamless CM production pipeline.

\textbf{Discussion.} It is worth mentioning that the number of interactions between customization and production should also be limited although cloud-assisted customization services can greatly improve user experience~\cite{FELS2017410}. In practice, there is a trade-off between customization interactions and product manufacturing. From the user experience perspective, the more customization interactions lead to a better user experience. However, the increased number of customization interactions may also increase the expenditure and prolong the production time of products, consequently affecting user experience. Therefore, it can enhance customization services to provide users with some well-designed product samples accompanied by tutorials in advance.

\section{Challenges and Advances}
\label{sec:challenges}
The convergence of AI technologies, IoT, SDN, and other new ICTs in smart factories, while it can significantly increase the flexibility, intelligence, and efficiency of CM systems it also poses challenges. In this section, we discuss these challenges as well as recent advances.

\subsection{Smarter devices in customized manufacturing}
In a CM environment, the equipment does not only perform the basic functions of automation but also needs to be intelligent and flexible. From the point of view of stand-alone equipment, the devices should have the following functions: parameter sensing, data storage, logical inference, information interaction, self-diagnostics, hybrid computing support, and preventive maintenance. From the perspective of the physical layer, the realization of CM is increasingly complex, as standard devices or equipment cannot accomplish this complicated task. Therefore, devices need to have both collaborative ability and swarm intelligence to collaborate with each to complete complex tasks. Devices with the above functions can meet the requirements of personalized production and adapt to the trend of achieving the automation and intelligence of manufacturing. Therefore, CM devices must be based on the integration and deep fusion of advanced manufacturing technologies, automation technologies, information technology, image technology, communication technologies, and AI technologies to realize the local and swarm intelligence. In this aspect, to be realizable in a generic context, further progress needs to be made in AI and in AI/robotics integration. However, as shown in Section~\ref{sec:case}, CM is already realizable in restricted environments.

Extensive studies have been conducted in the respective technologies to support CM~\cite{132,133,134}: intelligent connectivity, data-driven intelligence, cognitive Internet of things, and industrial big data. Unfortunately, most of these studies have mainly focused on single-device intelligence. Although these studies provide useful suggestions towards smarter devices, there is a still huge gap between realistic smart devices and the current solutions. We consider that an ideal method for realizing smart devices in manufacturing should integrate a series of methods to ensure the CM with edge intelligence that offers adaptability and response to a wide range of scenarios. These different methods or algorithms can be orchestrated by a central AI unit, which decides which algorithm is more useful for a particular task (combining a symbolic AI approach with machine learning). Meanwhile, a good balance between intelligence and cost is still a challenge that needs to be addressed in the future.

\subsection{Information interaction in CM}
Most manufacturing frameworks represent the distributed systems, where high-efficient communications between different components are needed~\cite{135,136}. On the one hand, different components of CM need not only effective connections but also highly efficient information interactions. As shown in early parts of this paper, the CM systems must incorporate different information interaction technologies and algorithms. On the other hand, due to the device heterogeneity and the increased demands of communications, the CM information interactive systems have to transmit massive data with different latency requirements. Consequently, the networks of CM need to be optimized to fulfill the diverse requirements of different applications, such as media access, moving handover, and congestion control. In other words, CM needs as a basis of a highly efficient information interaction system, which integrates multiple information technologies, including the absorbing DL and other AI algorithms. This system particularly depends on efficient and reasonable information interaction. With the development of manufacturing techniques, an increased number of studies have focused on the massive volume of communications and a large number of connections~\cite{137}. Based on network performance optimization methods, researchers have developed several industrial networks~\cite{138,139}. As different information transfer protocols exist with different requirements, such as time-sensitive and time-insensitive data transfer, the construction of a more efficient interaction industrial network represents another important challenge of future research. Existing studies mostly focus on high-speed communications, rather than efficient interaction with intelligent networks. These methods usually only meet the basic requirements of industrial networks, such as bandwidth and delay.

Recently, wireless mobile communications (e.g., 5G systems) and optical fiber communications have achieved great development~\cite{140,141}. The 5G technology for smart factories is still in its infancy, and there are still a number of issues such as deployment, accessing and spectrum management to be addressed. A hybrid industrial network with the latest communication technologies and AI (including deep learning, integrated learning, transfer learning, etc.) can be a promising solution for information interaction, which should consider different data flows according to different applications of intelligent CM.

\subsection{Dynamic reconfiguration of manufacturing resources}
Intelligent manufacturing, especially CM, involves a dynamic reorganization of the available resources and extreme flexibility. The essence of CM is to provide customized products, which have the characteristics of small-batch, short processing cycle, and flexible production. Therefore, manufacturing resources need constant readjustment and reorganization~\cite{142}. In addition, customized products are thought of as a variety of processing crafts. Moreover, the increasing complexity of the industrial environments makes the management of the resources even more difficult. A factory nowadays is a system consisting of many sub-systems, which can produce emergent behaviors through the interaction of the subsystems. Therefore, the dynamic reconstruction of manufacturing resources is one of the main challenges towards achieving a generic CM facility. In this aspect, a number of strategies have been proposed~\cite{143,144} in this domain.

Recently, knowledge reasoning, knowledge graph, transfer learning, and other AI algorithms have attained great progress~\cite{145,146,147}. In our opinion, hybrid AI methods combining the latest knowledge reasoning technologies with swarm intelligence can be a promising solution for dynamic reconstruction, which may consider different application scenarios of CM. Manufacturing-based process optimization with ML technologies may be one of the effective methods to reorganize resources. Moreover, digital twin technologies may be the driving technologies to improve resource reconfiguration~\cite{148,149,150}.

\subsection{Practical deployment and knowledge transfer}

Although the AI-assisted CM framework is promising to foster smart manufacturing, several challenges in industrial practice arise before the formal adoption of this framework. The first non-technical challenge mainly lies in the cost of upgrading existing manufacturing machinery, digitizing manufacturing equipment, and purchasing computing facilities as well as AI services. This huge cost may be affordable for small and medium-sized enterprises (SMEs). With respect to upgrading manufacturing production lines for SMEs, retrofitting legacy machines can be an economic solution as discussed in Section~\ref{subsec:edge-sensing}. In particular, diverse sensors and IoT nodes can be attached to existing manufacturing equipment to collect diverse manufacturing data. Those sensors and IoT nodes can be collected with the Internet so as to improve the interconnectivity of legacy machines. For example, Raspberry Pi models mounted with sensors can be deployed in workrooms to collect ambient data~\cite{YWang:TIA19,KZhang:TII20}. Besides hardware upgrading, software tools as well as AI services should be also purchased and adopted by manufacturing enterprises. Similarly, the economic solution for SMEs is to outsource manufacturing data to cloud services providers or Machine learning as a service (MLaaS) provider who can offer on-demand computing services. Nevertheless, outsourcing confidential data to untrusted third parties may increase the risks of security and privacy leakage. Thus, it is a prerequisite to enforce privacy and security protection schemes on manufacturing data before outsourcing. Moreover, the expenditure of system operating and training personnel should not be ignored in practical deployment. 

Another challenge is the effective technology transfer from research institutions to enterprises. Technology transfer involves many non-technical factors and multiple parties. The non-technical issues of technology transfer include marketing analysis, intellectual property management, technical invention protection, commercialization, and financial returns. Many frontier technological innovations often end up with unsuccessful technology transfer due to ignorance of the non-technical factors~\cite{JPan:IJPR19}. One of the main obstacles in technology transfer lies in the technology readiness level (TRL) gap between research and industrial practice. In particular, research institutions often focus on research results at TRL 1-3 implying basic feasibility and effectiveness while industrial enterprises often require transferred technologies at TRL 7-8 or even higher levels meaning prototype demonstration and real deployment~\cite{RYBICKA20161001}. We admit that there is still a long way for the AI-assisted CM framework before reaching TRL 7-8. The investigation of the technology transfer of AI-assisted CM will be a future direction. Both researchers and industrial practitioners are expected to work together to realize AI-assisted CM.

\section{Conclusion}
\label{sec:conc}
Recent advances in AI technologies have had an impact on the manufacturing industry, especially within customized smart manufacturing. In this article, AI-assisted customized manufacturing architectures – incorporating IoT, edge intelligence, and cloud computing paradigms – have been proposed. These key AI-enabled technologies have been validated in an industrial packaging scenario. Further, each of the aspects composing these architectures have been carefully reviewed.

The fusion of AI and manufacturing provides a potential solution for customized manufacturing. Future research will be directed towards tackling the challenges related to smart manufacturing devices, effective information interaction, dynamic
reconstruction of manufacturing resources, and practical deployment issues.


%

\appendices



\ifCLASSOPTIONcaptionsoff
  \newpage
\fi



\bibliographystyle{IEEEtran}
\bibliography{mylib}
%

%

\begin{IEEEbiography}[{\includegraphics[width=1in,height=1.25in,clip,keepaspectratio]{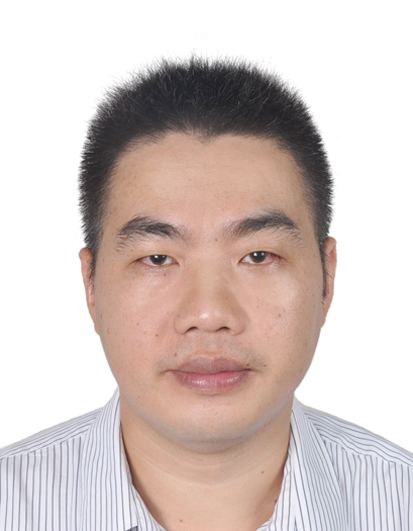}}]{Jiafu Wan}(M'11) is a Professor in School of Mechanical \& Automotive Engineering at South China University of Technology, China. He has directed 20 research projects, including the National Key Research and Development Project, and the Joint Fund of the National Natural Science Foundation of China and Guangdong Province. Thus far, he has published more than 160 scientific papers, including 100+ SCI-indexed papers, 50+ IEEE Trans./Journal papers, 20 ESI Highly Cited Papers and 4 ESI Hot Papers. According to Google Scholar Citations, his published work has been cited more than 12,000 times (H-index = 49). His SCI other citations, sum of times cited without self-citations, reached 3500 times (H-index = 36) according to Web of Science Core Collection. He is an Associate Editor of IEEE/ASME Transactions on Mechatronics, Journal of Intelligent Manufacturing and Computers \& Electrical Engineering, and Editorial Board for Computer Integrated Manufacturing Systems. He is a Leading Guest Editor for several SCI-indexed journals, such as IEEE Systems Journal, IEEE Access, Elsevier Computer Networks, Mobile Networks \& Applications, Computers and Electrical Engineering, Wireless Communications and Mobile Computing, Journal of Internet Technology, and Microprocessors and Microsystems. His research interests include Cyber-Physical Systems, Intelligent Manufacturing, Big Data Analytics, Industry 4.0, Smart Factory, and Cloud Robotics. He is listed as a Clarivate Analytics Highly Cited Researcher in 2019 and 2020.
\end{IEEEbiography}
\begin{IEEEbiography}[{\includegraphics[width=1in,height=1.25in,clip,keepaspectratio]{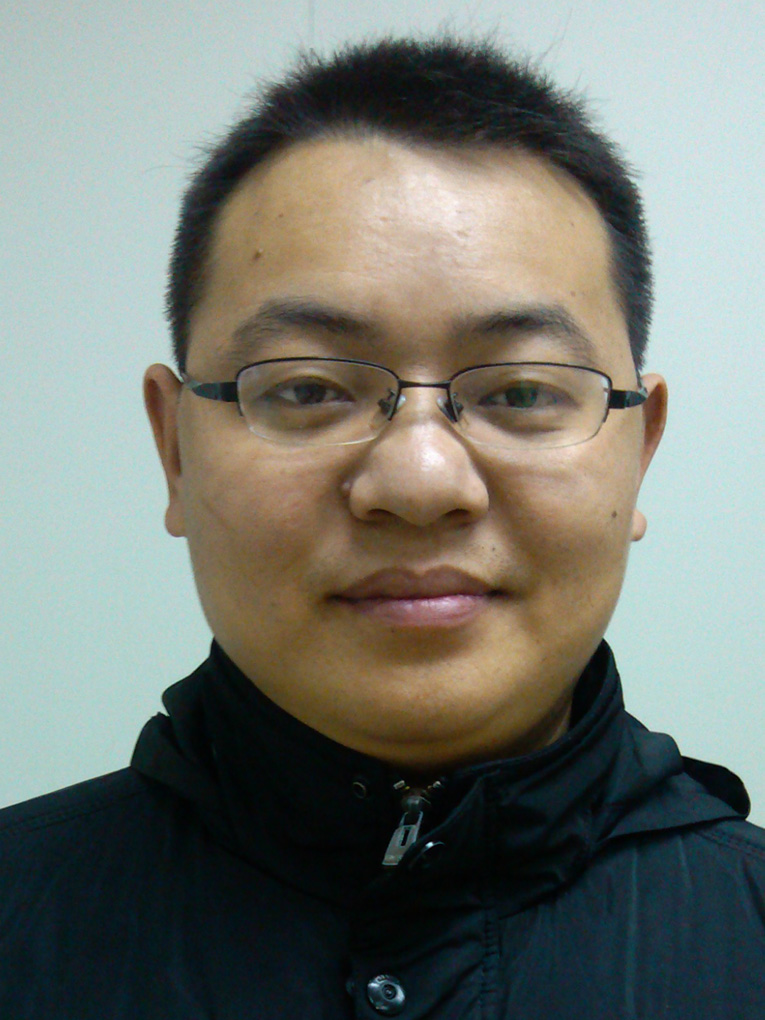}}]{Xiaomin Li}
is currently an Associate Professor at Zhongkai University of Agriculture and Engineering, Guangzhou, China. He received B.S. degree from the Air Force Engineering University, Xi'an China (2004); M.S. degree from the South China Agricultural University, Guangzhou, China (2014); and Ph.D. degree from the South China University of Technology, Guangzhou, China (2018). His research interests include Cyber-Physical Systems, Smart manufacturing, Big data, Wireless Network.
\end{IEEEbiography}
\begin{IEEEbiography}[{\includegraphics[width=1in,height=1.25in,clip,keepaspectratio]{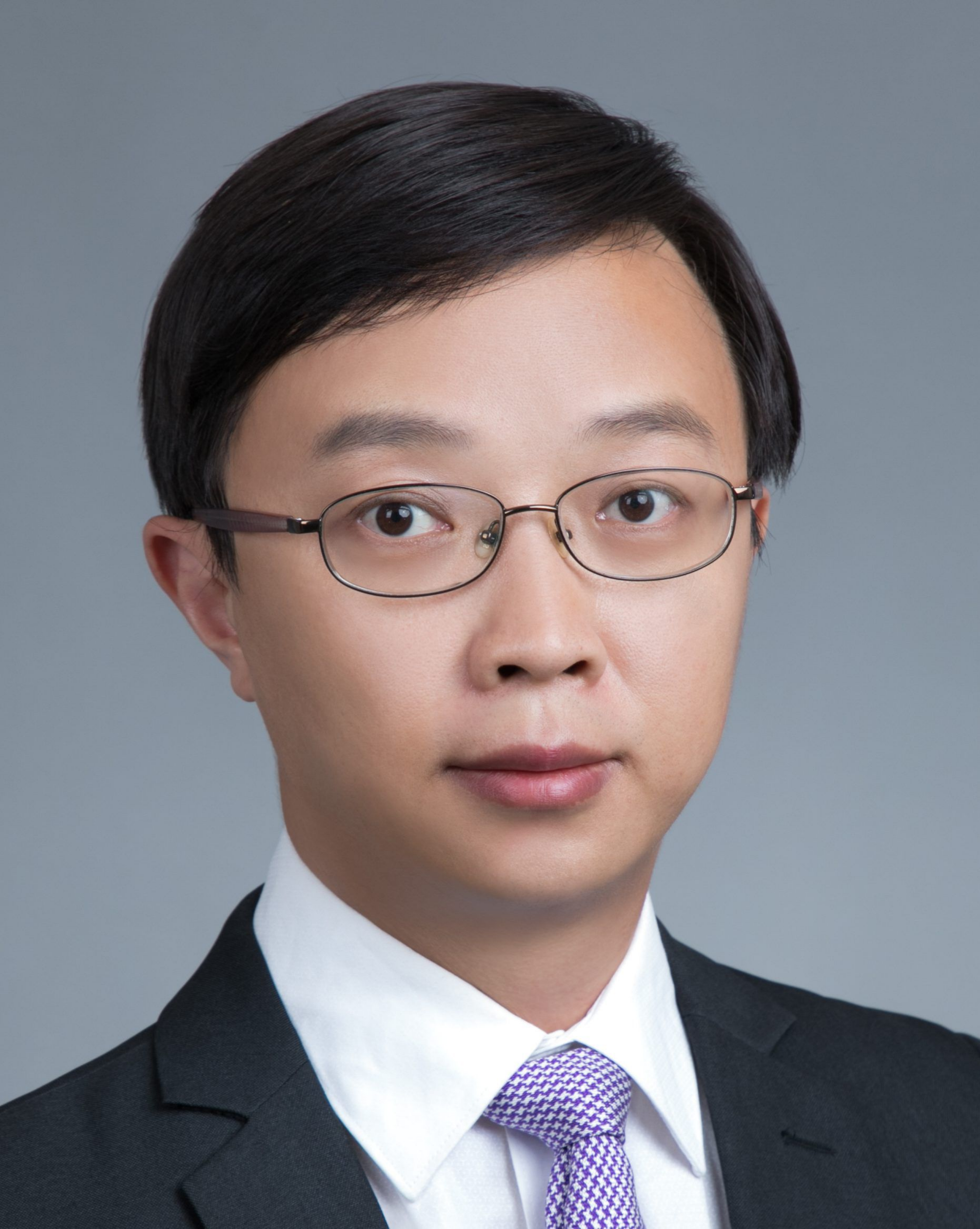}}]{Hong-Ning Dai}(M’05–SM’16) is currently with Faculty of Information Technology at Macau University of Science and Technology (MUST) as an associate professor. He obtained the Ph.D. degree in Computer Science and Engineering from Department of Computer Science and Engineering at the Chinese University of Hong Kong. His current research interests include Data Analytics, Blockchain Technology, Cyber-Physical System and industrial Internet of Things. He is a holder of 1 U.S. patent and 1 Australia innovation patent. He is also the winner of Bank of China (BOC) Excellent Research Award of MUST. He has 3 ESI highly-cited papers awarded by Clarivate Analytics and 1 IEEE Outstanding Paper Award conferred by IEEE International Conference on Cyber Physical and Social Computing. He holds visiting positions at Department of Computer Science and Engineering of the Hong Kong University of Science and Technology, School of Electrical Engineering and Telecommunications of the University of New South Wales, Hong Kong Applied Science and Technology Research Institute (ASTRI). He has served as associate editors of IEEE Systems Journal, IEEE Access, and Connection Science, an editor of Ad Hoc Networks, guest editors for IEEE Transactions on Industrial Informatics, IEEE Transactions on Emerging Topics in Computing. He is also an ACM senior member. 
\end{IEEEbiography}
\begin{IEEEbiography}[{\includegraphics[width=1in,height=1.25in,clip,keepaspectratio]{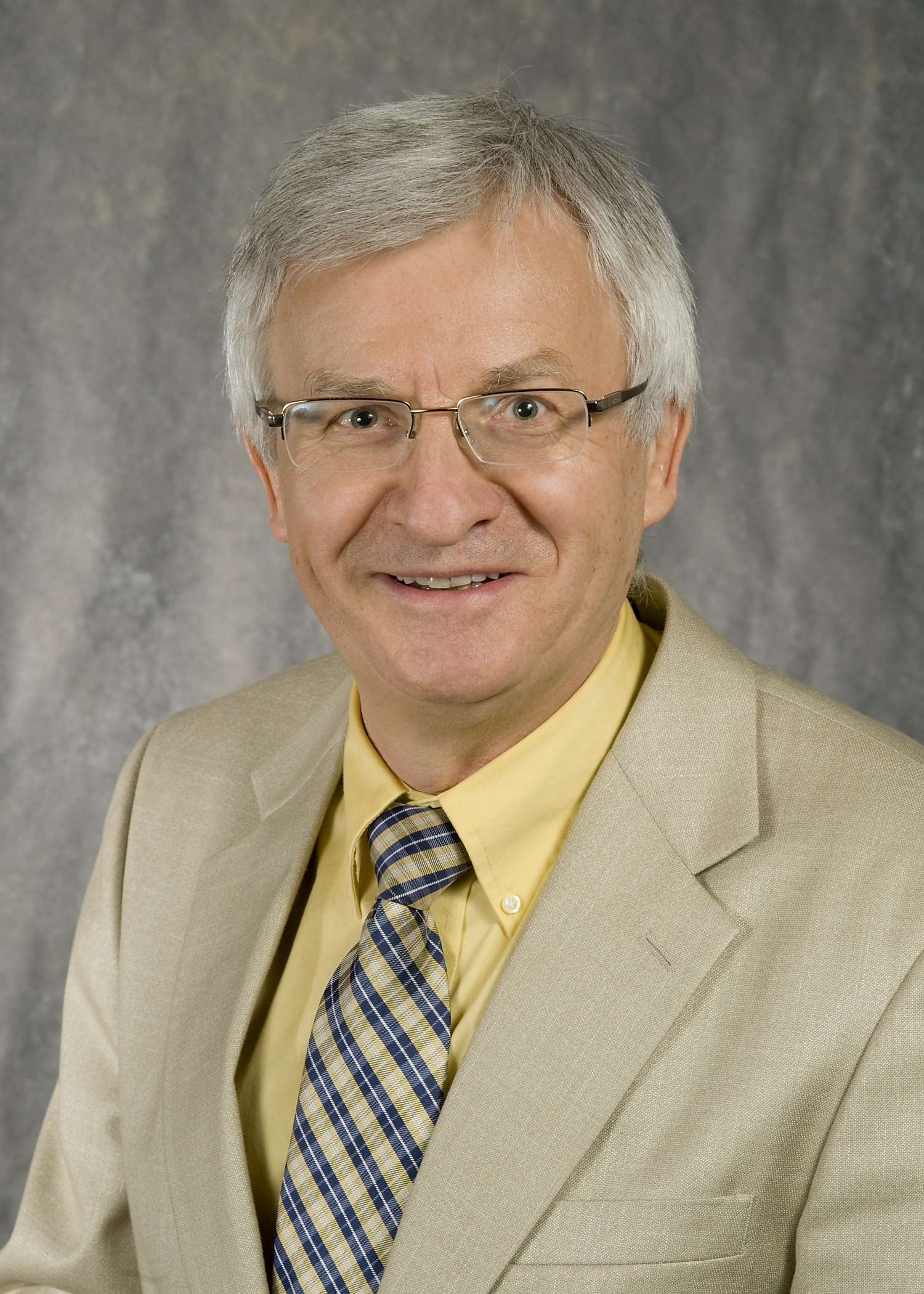}}]{Andrew Kusiak}
is a Professor in the Department of Industrial and Systems Engineering and Director of the Intelligent Systems Laboratory at The University of Iowa, Iowa City. He has chaired two departments, Industrial Engineering (1988-95) and Mechanical and Industrial Engineering (2010-15). His current research focuses on applications of artificial intelligence and big data in smart manufacturing, product development, renewable energy, sustainability, and healthcare. He has published numerous books and technical papers in journals sponsored by professional societies, such as the Association for the Advancement of Artificial Intelligence, the American Society of Mechanical Engineers, Institute of Industrial and Systems Engineers, Institute of Electrical and Electronics Engineers, Nature, and other societies. He speaks frequently at international meetings, conducts professional seminars, and consults for industrial corporations. Dr. Kusiak has served in elected professional society positions as well as various editorial roles in over fifty journals, including five different IEEE Transactions. Professor Kusiak is a Fellow of the Institute of Industrial and Systems Engineers and the Editor-in-Chief of the Journal of Intelligent Manufacturing.
\end{IEEEbiography}
\begin{IEEEbiography}[{\includegraphics[width=1in,height=1.25in,clip,keepaspectratio]{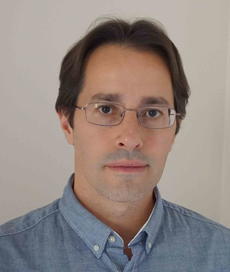}}]{Miguel Mart\'{i}nez-Garc\'{i}a} is a Lecturer in Humanmachine systems at Loughborough University, UK. He received a BSc degree in Mathematics from the Polytechnic University of Catalonia (UPC), Spain in 2013, a MSc in Advanced Mathematics and Mathematical Engineering (MAMME) from the same university in 2014, and a PhD in Engineering from the University of Lincoln, UK, in 2018. He also worked as a researcher since 2017, both at the University of Lincoln and at the Advanced Virtual Reality Research Centre (AVRRC) at Loughborough University. His research interests include human-machine integration, machine learning, artificial intelligence, intelligent signal processing, and complex systems, with particular focus in the analysis of non-linear signals representing phenomena of interest between humans and machines.
\end{IEEEbiography}
\begin{IEEEbiography}[{\includegraphics[width=1in,height=1.25in,clip,keepaspectratio]{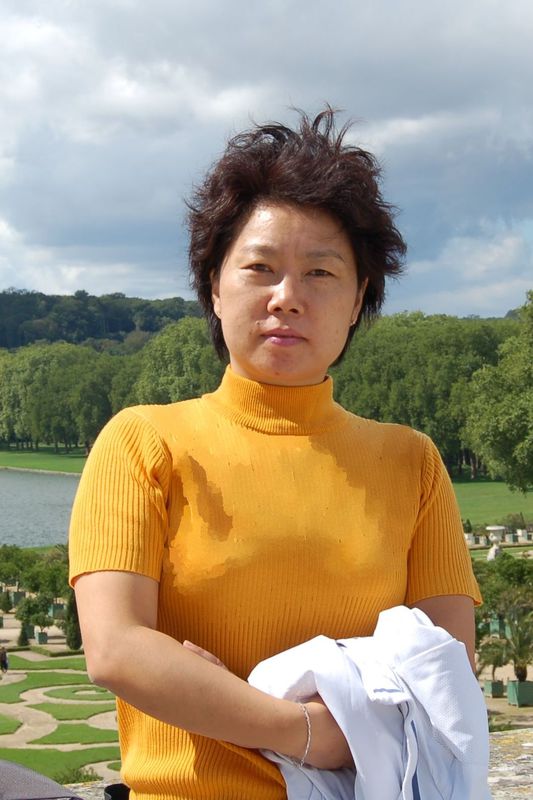}}]{Di Li}
is a Professor of School of Mechanical and Automotive Engineering, South China University of Technology, Guangzhou, China. Di Li was born in Fu yang City, Anhui Province, China, in 1965. She received the B.S. and M. S. degrees in aerospace engineering from the Nanjing University of Aeronautics and Astronautics, Nanjing City, Jiangshu Province, China, in 1988 and the Ph.D. degree in automation from the South China University of Technology, Guangzhou City, Guangdong Province, China, in 1993. From 1988 to 1990, she was an Assistant Engineer with the Qingdao Beihai Shipyard of the China State Shipbuilding Corporation, Qingdao City, Shandong Province, China. From 1993 to 1998, she was an Associate Professor with the Guangdong University of Technology, Guangzhou City, Guangdong Province, China. Since 1999, she has been a Professor with the Mechanical and Electrical Engineering Department, South China University of Technology. She is the author of more than 150 articles and holds eight patents. Her research interests include high-performance embedded control system, numerical control, and machine vision. Ms. Li was a recipient of the First Prize for Science and Technology Development of Guangdong Province in 2009 and Ding Ying Science and Technology Awards in 2011.
\end{IEEEbiography}



\end{document}